\documentclass[10pt,twocolumn,letterpaper]{article}

\usepackage[pagenumbers]{cvpr} 
\usepackage{svg}
\definecolor{cvprblue}{rgb}{0.21,0.49,0.74}
\usepackage[pagebackref,breaklinks,colorlinks,allcolors=cvprblue]{hyperref}
\usepackage{makecell}
\usepackage{multirow}
\usepackage{float}
\usepackage{amssymb}

\def\paperID{10254} 
\def\confName{ICCV}
\def\confYear{2025}

\title{IQA-Adapter: Exploring Knowledge Transfer from Image Quality Assessment to Diffusion-based Generative Models}
\author {
    Abud Khaled\textsuperscript{\rm 1,3},
    Sergey Lavrushkin\textsuperscript{\rm 1,2},
    Alexey Kirillov\textsuperscript{\rm 3,4},
    Dmitriy Vatolin\textsuperscript{\rm 1,2,3}\\
    \textsuperscript{\rm 1}MSU Institute for Artificial Intelligence\\
    \textsuperscript{\rm 2}ISP RAS Research Center for Trusted Artificial Intelligence\\
    \textsuperscript{\rm 3}Lomonosov Moscow State University\\
    \textsuperscript{\rm 4}Yandex\\
    \{khaled.abud, sergey.lavrushkin, alexey.kirillov, dmitriy\}@graphics.cs.msu.ru
}

\begin{document}
\maketitle
\begin{abstract}
Diffusion-based models have recently revolutionized image generation, achieving unprecedented levels of fidelity. However, consistent generation of high-quality images remains challenging partly due to the lack of conditioning mechanisms for perceptual quality. In this work, we propose methods to integrate image quality assessment (IQA) models into diffusion-based generators, enabling quality-aware image generation. We show that diffusion models can learn complex qualitative relationships from both IQA models’ outputs and internal activations. 
First, we experiment with gradient-based guidance to optimize image quality directly and show this method has limited generalizability. To address this, we introduce \textbf{IQA-Adapter}, a novel framework that conditions generation on target quality levels by learning the implicit relationship between images and quality scores. 
When conditioned on high target quality, IQA-Adapter can shift the distribution of generated images towards a higher-quality subdomain, and, inversely, it can be used as a degradation model, generating progressively more distorted images when provided with a lower-quality signal. Under high-quality condition, IQA-Adapter achieves up to a 10\% improvement across multiple objective metrics, as confirmed by a user preference study, while preserving generative diversity and content. Furthermore, we extend IQA-Adapter to a reference-based conditioning scenario, utilizing the rich activation space of IQA models to transfer highly specific, content-agnostic qualitative features between images.
\end{abstract}    
\section{Introduction}
\label{sec:intro}
\begin{figure*}[ht!]
\centering
   \includegraphics[width=.77\textwidth]{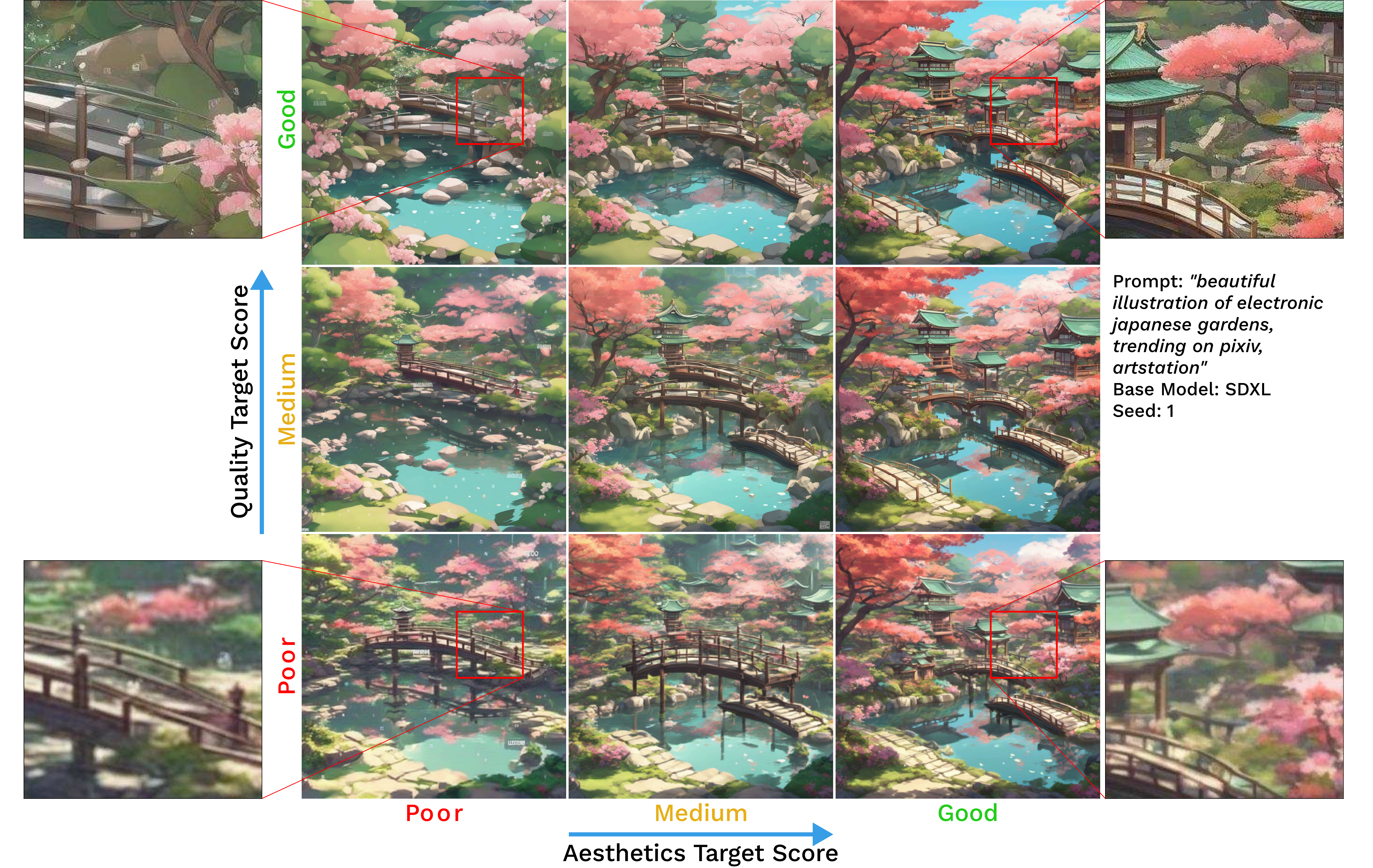}
  \caption{Quality-aware image generation with \textbf{IQA-Adapter}. All images are generated with the SDXL base model, the same prompt, and the seed. The IQA-Adapter is trained with TOPIQ \cite{topiq} and LAION-Aesthetics \cite{laion_aes} metrics.}
  \label{fig:demo_2d}%
\vspace*{-0.4cm}
\end{figure*}

Recent advances in diffusion-based models have greatly improved text-to-image generation, producing highly realistic visuals from textual prompts. Models like DALL-E 3 \cite{betker2023improving}, FLUX \cite{flux}, and SDXL \cite{podellsdxl} exemplify this progress. Moreover, recent extensions leverage diverse guidance sources such as depth, pose \cite{zhang2023adding}, or reference images \cite{ye2023ip-adapter}, enhancing control and flexibility. Unified frameworks integrating multiple conditioning types, such as OmniGen \cite{xiao2024omnigen}, have recently emerged to further extend generation capabilities.

Despite these improvements, conditioning generative models explicitly on image quality or aesthetics from Image Quality and Aesthetic Assessment (IQA/IAA) systems remains largely unexplored. IQA methods evaluate images according to human-perceived quality, while IAA focuses on more subjective, content-dependent aspects. Integration of IQA/IAA models directly into the generative architectures is a logical next step towards aligning generated images with human preferences, moving beyond traditional text-image alignment metrics. Although recent works \cite{xu2024imagereward, Kirstain2023PickaPicAO, wu2023better} began exploring aesthetic alignment, explicit incorporation of IQA/IAA knowledge into generative models has not yet been systematically addressed.

Motivated by recent successful transfers of generative priors to IQA \cite{fu2024dp, de2024genziqa, li2024feature, wang2024diffusion}, we propose the opposite direction: incorporating IQA expertise into generative diffusion models via conditioning. Specifically, we present \textbf{IQA-Adapter}, a novel conditioning architecture leveraging IQA scores, enabling generation of images with controlled levels of image quality and aesthetics.

We summarize our contributions as follows:
\begin{itemize}
\item \textbf{Qualitative adaptation method.} We introduce IQA-Adapter, a novel conditioning tool for diffusion models that enables quality-aware generation guided by IQA/IAA scores. To our knowledge, this work is the first systematic attempt to directly introduce IQA knowledge in a generative setting via conditioning: we show that diffusion models can implicitly learn complex relationships from both IQA models’ outputs and internal activations.
\item \textbf{Diverse IQA/IAA model integration.} We experiment with a range of IQA and IAA models with diverse architectures and training datasets, demonstrating the adaptability of our approach to different quality and aesthetic metrics and the generalization of quality features learned by IQA-Adapter. Furthermore, we employ a gradient-based quality optimization method to explore adversarial patterns that emerge within images generated with a high IQA-guidance scale.
\item \textbf{Reference-based conditioning.} We adopt IQA-Adapter for qualitative image-prompting scenario, demonstrating that IQA models can be used to transfer highly specific, content-agnostic qualitative features from the reference to a generated image.
\end{itemize}

\section{Related Work}
\label{sec:related_work}

\textbf{Generative Models.} Diffusion models recently set new standards in image generation. Early diffusion-based methods~\cite{ramesh2021zero, rombach2022high, saharia2022photorealistic} showed significant gains over previous GAN- and VAE-based approaches. Later advances~\cite{liu2024playground, kastryulin2024yaart, dai2023emu, baldridge2024imagen, esser2403scaling, betker2023improving, flux, chen2024pixart} further improved visual quality, aesthetics, and relevance via better data, larger models, architectural refinements, and alternative diffusion architectures.
Our work continues this line of research by bridging the gap between image generation and quality assessment tasks.

\textbf{Adapters and Customization.} Recent works introduced diverse adapters for finer control and personalization. LoRA~\cite{hu2022lora} offered efficient fine-tuning via low-rank decomposition. Dreambooth~\cite{ruiz2023dreambooth}, Textual Inversion~\cite{galimage}, and IP-Adapter~\cite{ye2023ip-adapter} enabled user-specific generation. ControlNet~\cite{zhang2023adding}, T2IAdapter~\cite{mou2024t2i}, ConceptSliders~\cite{gandikota2023erasing} added spatial or attribute-specific guidance, while StyleCrafter~\cite{liu2023stylecrafter} focused on style-transfer task. Unlike existing methods conditioned on text, images, or masks, our adapters uniquely condition generation on numerical values encoding continuous semantic attributes (e.g., aesthetics or quality).

Some studies~\cite{polyak2024movie, alonso2024diffusion, podellsdxl} condition generation on technical attributes (e.g., resolution, crops~\cite{podellsdxl}). Differently, our approach conditions on high-level semantic features obtained automatically from pretrained IQA models.

\textbf{IQA and IAA.} Image Quality Assessment (IQA) methods quantify technical degradations (e.g., artifacts) using full- or no-reference techniques, while Image Aesthetic Assessment (IAA) evaluates subjective visual aspects (composition, color harmony, aesthetics).

Earlier IQA/IAA methods utilized handcrafted features modeling human perception explicitly~\cite{ssim, ms_ssim, vif, brisque, niqe, diivine, ilniqe}. Modern approaches rely on deep neural networks trained on annotated datasets~\cite{hyperiqa, musiq, topiq, dbcnn, arniqa, liqe, tres, yang2022maniqa, cnniqa, clipiqa, nima, hosu2020koniq, jinjin2020pipal, spaq, clive, flive, ava, kadid, aadb}. Recent works integrated generative priors and diffusion models to improve IQA/IAA metrics further~\cite{fu2024dp, de2024genziqa, li2024feature, wang2024diffusion}. Unlike prior studies transferring generative knowledge into IQA tasks, we uniquely integrate IQA knowledge into diffusion architectures for quality-aware image generation.

\begin{figure*}[ht!]
\centering
   \includegraphics[width=.8\textwidth]{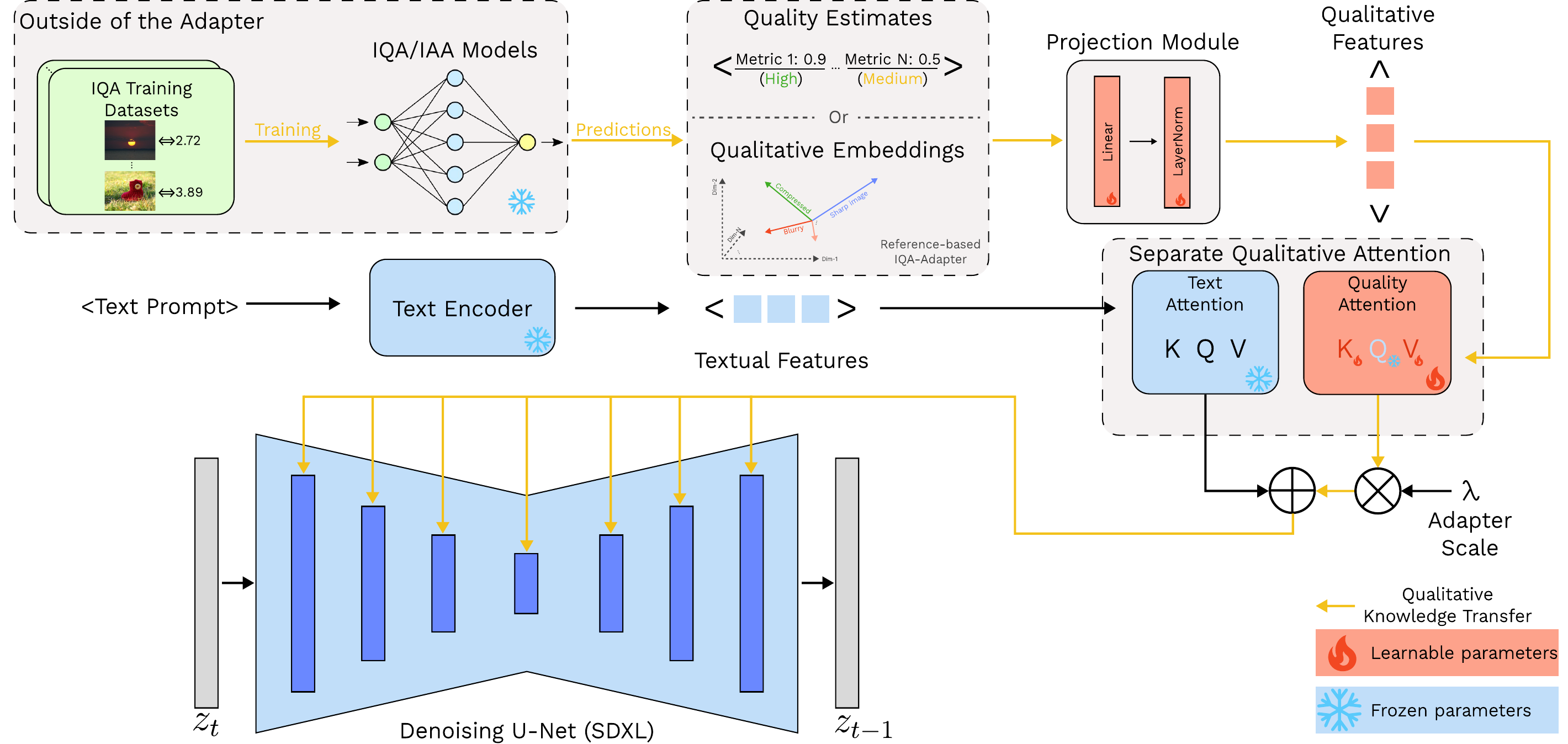}
  \caption{Overall architecture of the proposed \textbf{IQA-Adapter}. Yellow arrows depict IQA/IAA knowledge flow into the diffusion-based generator.}
  \label{fig:arch}
\vspace*{-0.4cm}
\end{figure*}

\textbf{Generation Quality Improvement.} 
Several recent works~\cite{diffusion_dpo, q_refine, vmix, beautifulprompt, compel} have introduced various approaches for improving generation quality. Some focus on automatic~\cite{beautifulprompt} or manual~\cite{compel} prompt enhancement, while others rely on handcrafted high-quality datasets~\cite{vmix}. DiffusionDPO~\cite{diffusion_dpo} employs an aesthetic critic model to fine-tune the generator using reinforcement learning approaches to maximize quality. Q-Refine~\cite{q_refine} utilizes an IQA model to detect low-quality regions in the image and inpaints them using off-the-shelf image enhancement models without any qualitative knowledge transfer to the generator. In contrast to prior work, we do not focus solely on maximizing the target quality score, but empower the generative model with the ability to modulate its output across a wide qualitative spectrum. To this end, we employ rich qualitative priors of pretrained IQA and IAA models and transfer their knowledge directly to the generator.
\section{Learning the relationship between images and visual quality from IQA models}
\subsection{Baselines}

To establish a baseline for integration of IQA model knowledge into the generation process, we introduce a technique inspired by classifier guidance \cite{classguidance}. In our adaptation, we leverage NR-IQA models rather than a classifier, interpreting IQA scores as soft probabilities that reflect the likelihood of an image achieving high perceptual quality. This approach uses feedback from the IQA model to iteratively optimize image quality during the generation process:
\begin{equation*}
\hat{\epsilon}_\theta (z_{t-1}|c_t,f_{\phi})=\epsilon_{\theta}(z_t|c_t) + \alpha \cdot \omega(t)\nabla_{z_t} \log f_{\phi}(D(z_t)),
\end{equation*}
where $\epsilon_{\theta}$ is a latent diffusion model, $c_t$ is a textual condition, $f_{\phi}$ is a NR metric, $z_t$ represents the latent image at the $t$-th diffusion step, and $D(\cdot)$ is the VAE’s decoder that maps the latent representation back to image space. The parameter $\alpha$ allows adjustment of the IQA guidance weight, balancing the impact of quality conditioning, while the scaling coefficient $\omega(t)$ modulates the gradient's influence over time, linearly increasing from 0 to 1.

Although this method optimizes the target IQA score, its reliance on gradient-based adjustments introduces the risk of exploiting vulnerabilities within the IQA model. This can result in images that receive high ratings from the IQA model yet exhibit noticeable visual distortions --- a phenomenon similar to adversarial attacks, which we further discuss in Section \ref{sec:disc_adv_patterns}. 


\subsection{IQA-Adapter}

To address limitations of inference-time gradient optimization, we propose a method that implicitly learns a relationship between images and their corresponding quality assessments. 
By learning this connection, the generative model can internalize features associated with target-quality images and avoid characteristics linked to opposite quality. For instance, when conditioned on high-quality parameters, the model should generate images with fine-grained details and vibrant colors. Conversely, when conditioned on low quality, it should reproduce artifacts such as JPEG compression distortions or blurring.

\subsubsection{Architecture}
\label{sec:arch}
To condition the generative model on image quality, we leverage an adapter-based approach. It is a common concept for diffusion model customisation and is widely used in various tasks \cite{ye2023ip-adapter, mou2023t2i, liu2023stylecrafter}. The core idea of the approach is to project new data into additional tokens, which are then integrated into the model via cross-attention mechanisms, enabling the base model to receive detailed conditioning information from the new sources without altering its core weights. We selected adapter-based architecture for its lightweight design, ability to preserve core model's weights, and relatively small overhead during training and inference (Sec. \ref{sec:comp_overhead}).

Figure \ref{fig:arch} demonstrates the overall structure of our quality-conditioning framework, which we name \textbf{IQA-Adapter}. In this setup, quality scores are projected into tokens matching the dimensionality of textual tokens through a small projection module, consisting of a linear layer and LayerNorm \cite{ba2016layernormalization}. These tokens then enter the main generative model (U-Net in the case of SDXL \cite{podellsdxl}) via cross-attention layers. This design allows the diffusion model to modulate image quality based on target IQA scores. We further report our architectural experiments in the Ablation Study in Supplementary Sec. \ref{sec:abl_study}.

IQA-Adapter can accept multiple IQA scores as input, allowing for the integration of various IQA/IAA models that capture different aspects of image fidelity, e.g., quality in terms of distortions and overall aesthetics. To ensure consistency, all metric values are standardized to have zero mean and unit variance based on the training dataset.

\textbf{Separate Qualitative Attention}. Disentanglement of the qualitative information from the contextual condition provided by the textual prompt is an important feature of IQA-Adapter. It is done with a separate Qualitative Attention mechanism, which processes adapter tokens independently from the textual ones. Specifically, the adapter adds an additional cross-attention layer for each existing cross-attention operation in the base model. Without an IQA-Adapter, the base model processes the textual conditioning $c_t$ as follows:
\vspace{-0.2cm}
\begin{equation*}
\vspace{-0.1cm}
\operatorname{CrossAttn}(Z, c_t) = \operatorname{Softmax}\left(\frac{Q K^T}{\sqrt{d}}\right) V
\end{equation*}
\noindent where $Q = Z W_q$, $K = c_t W_k$, $V = c_t W_v$; $Z$ are image features, $d$ is the projection space dimension. With the IQA-Adapter, the attention mechanism is modified as follows:
\vspace{-0.15cm}
\begin{multline*}
\vspace{-0.1cm}
\operatorname{CrossAttn}(Z, c_t, c_q) = \\
\operatorname{Softmax}\left(\frac{Q K^T}{\sqrt{d}}\right) V + \lambda\times\operatorname{Softmax}\left(\frac{Q K'^T}{\sqrt{d}}\right) V'
\end{multline*}
\noindent where $K' = c_q W'_k$, $V' = c_q W'_v$, and $c_q$ are the quality conditioning features. Notably, the query matrix $W_q$ that processes the generated image features $Z$ is shared across both attention operations. This setup allows the IQA-Adapter to learn and apply quality-specific attributes in a content-agnostic way and generalize them across various textual contexts. To control the strength of the IQA-Adapter during inference, we introduce a scaling parameter $\lambda$, which adjusts the impact of quality conditioning by modifying the cross-attention term for quality features.
\begin{figure*}[ht!]
\centering
   \includegraphics[width=.98\textwidth]{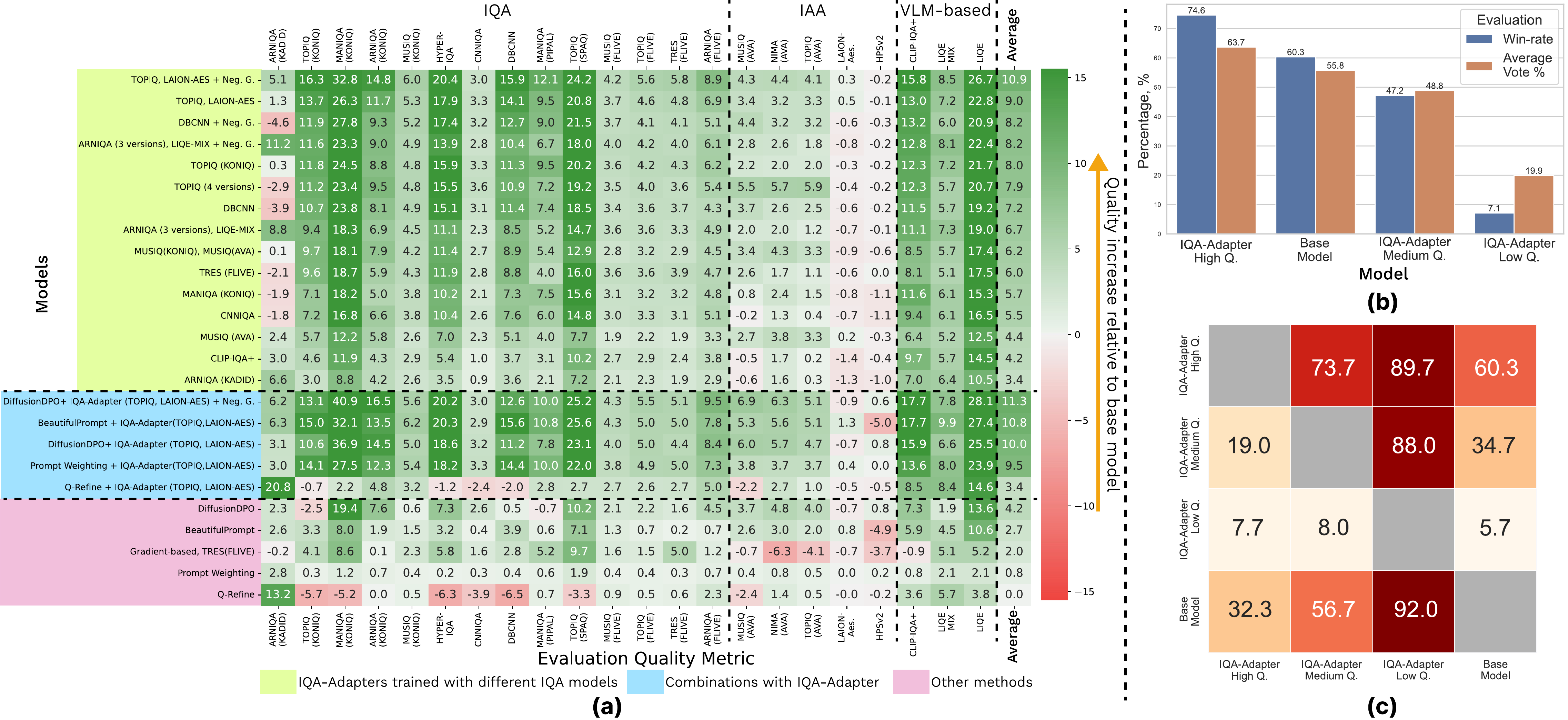}
  \caption{(a) Quality improvement relative to the base model (in \%) for the IQA-Adapters trained on different IQA/IAA models. All IQA-Adapters are conditioned with high target quality (99th percentile of the training dataset) and use the same prompts and seeds. "+Neg. G." denotes qualitative negative guidance. Prompts are taken from Lexica.art user-generated prompts dataset. (b,c) Results of the side-by-side subjective study of the IQA-Adapter conditioned on different quality levels. (b) Overall results of all comparisons. (c) Pair-wise win rates. }
  \label{fig:heatmap_adapter}%
\vspace*{-0.35cm}
\end{figure*}

\textbf{Qualitative Negative Guidance}. Since the concept of visual quality has clearly defined notions of "good" and "bad," it becomes feasible to adopt \textit{negative guidance}, akin to its application in text-based generation. For textual conditioning, it involves using an additional prompt that is semantically opposite to the main one in the unconditional part of the classifier-free guidance. It pushes the latent representation of the image away from producing undesired features. To enable qualitative negative guidance, we modify the classifier-free guidance mechanism as follows:
\begin{multline*}
    \hat{\epsilon}_\theta (z_{t}|c_t, q) 
    = \epsilon_{\theta}(z_t| c_t^{\text{neg}}, q^{\text{neg}})
    +  \\
    + g \cdot \big(
        \epsilon_{\theta}(z_t| c_t, q)
        - \epsilon_{\theta}(z_t| c_t^{\text{neg}}, q^{\text{neg}})
    \big)
\end{multline*}
where $\epsilon_{\theta}$ is a latent diffusion model, $g$ is guidance scale, $c_t$ and $c_t^{\text{neg}}$ are positive and negative textual prompts, $q$ is a desired qualitative condition, and $q^{\text{neg}}$ specifies an opposite quality level. Since the input scores are normalized with a mean of 0, we can set $q^{\text{neg}} = - \delta \cdot q$, where parameter $\delta$ controls the “gap” between the opposite quality levels, modulating the strength of the negative guidance. This optional step can be used during inference to boost the adapter's effect, even with moderate adapter scales; however, when used with excessively large scale, it can cause undesired “over-stylisation” effects (Sec. \ref{sec:adv_patterns_supp}).

\subsection{Reference-based IQA-Adapter}
In addition to direct conditioning on the target quality level, we have also explored the scenario of qualitative transfer from an existing reference image. For this purpose, we condition a generative model on the activations extracted from the  intermediate layers of an IQA model. More specifically, we apply a pretrained IQA model to the reference image to extract a qualitative embedding, which we then pass through the projection module to obtain qualitative tokens for the subsequent attention operation. We name this variation of the method \textit{Reference-based IQA-Adapter}.

In the conditioning process, we exploit a useful property of activations of some IQA models (especially those from the farthest layers): they contain almost no information about the semantics of the image, but accumulate information necessary for quality assessment (e.g., type and strength of distortions). This allows us to extract mostly qualitative, content-agnostic knowledge from the reference image, preventing “leakage” of unwanted information (e.g. objects, faces, colors). In particular, we used the ARNIQA \cite{arniqa} IQA model to obtain qualitative embeddings, whose authors purposefully achieve this property of the activation space using a special training procedure. Our experiments (Sec. \ref{sec:ref_exps}) also confirm that it is well-suited for a reference-based scenario.

\subsection{IQA-Adapter Training}
\label{sec:adapter_training}
We train IQA-Adapter on triplets (\text{image}, \text{text}, \text{input quality scores}) where the image-text pairs are drawn from a text-to-image dataset, and the quality scores are estimated by passing each image through a target IQA/IAA model. The training follows the standard denoising diffusion probabilistic model (DDPM) procedure \cite{ho2020denoising}. In this process, a random timestep \(t \sim U[0, 1]\) is sampled, and noise is incrementally applied to the image  $x$ at the corresponding noise scale.
The model then learns to predict the added noise with the following objective:
\begin{equation*}
\mathcal{L} = \mathbb{E}_{x, t, \epsilon} \left[\|\epsilon - \epsilon_\theta(x_t | c_t, c_q)\|^2\right]
\end{equation*}
where $x_t$ is a noised representation of the input image, \(c_t\) is the textual condition, $c_q$ is the qualitative condition, $\epsilon$ is the added noise, and $\epsilon_\theta(x_t | c_t, c_q)$ is the predicted noise. 

During this process, only the adapter weights are adjusted to allow the generative model for incorporating quality score information and steer the output generation accordingly. To maintain flexibility for classifier-free guidance during inference, we randomly drop the textual and quality conditions with a small probability, which encourages the model to generate images unconditionally.

A key advantage of IQA-Adapter is that it does not require backpropagation through the IQA models (as it only uses quality scores of training images), enabling the use of non-differentiable metrics or even ground-truth subjective scores from sufficiently large subjective studies. As demonstrated in Section \ref{sec:hq_cond}, bypassing gradient-based optimization significantly improves the robustness of the method and its transferability across metrics beyond those used for training, enhancing the generality of the learned quality features across various evaluation models.

The training of the Reference-based IQA-Adapter is fairly similar: a qualitative embedding is obtained from the image being reconstructed, which is further used as an additional condition for image denoising. Notably, in order to expand the coverage of the IQA model’s activation space during adapter training, we also employed an image degradation model introduced in \cite{arniqa} as an additional augmentation with a small probability ($p=0.1$).
\section{Experiments and Evaluation}
\label{sec:exp_eval}
\subsection{Experimental Setup}

\textbf{Models.} For all experiments involving both gradient-based guidance and IQA-Adapter, we used SDXL as the base model. The IQA-Adapters were trained on the CC3M \cite{sharma2018conceptual} dataset ($\sim$3 million images) for 24,000 steps, followed by fine-tuning on a subset \cite{laion_subset} of the LAION-5B \cite{laion5b} dataset ($\sim$170k images with an aesthetics score $>$ 6.5) for an additional 3,000 steps. Training a single IQA-Adapter model required approximately 260 Nvidia A100 80GB GPU hours. We employed two tokens for qualitative features. For more details on the IQA-Adapter training, refer to the Supplementary Section \ref{sec:adapter_training_supp}. 
The code for training and inference, as well as pre-trained weights for IQA-Adapters, will 
be available in our GitHub repository: 
\url{https://github.com/X1716/IQA-Adapter}.

During inference, we used a guidance scale of 7.5 and 35 sampling steps. For the IQA-Adapters, we set the adapter scale to $\lambda=0.5$, while for the gradient-based method, we applied a quality-guidance scale of $\alpha=30$.


\begin{figure*}[ht!]
\centering
   \includegraphics[width=.9\textwidth]{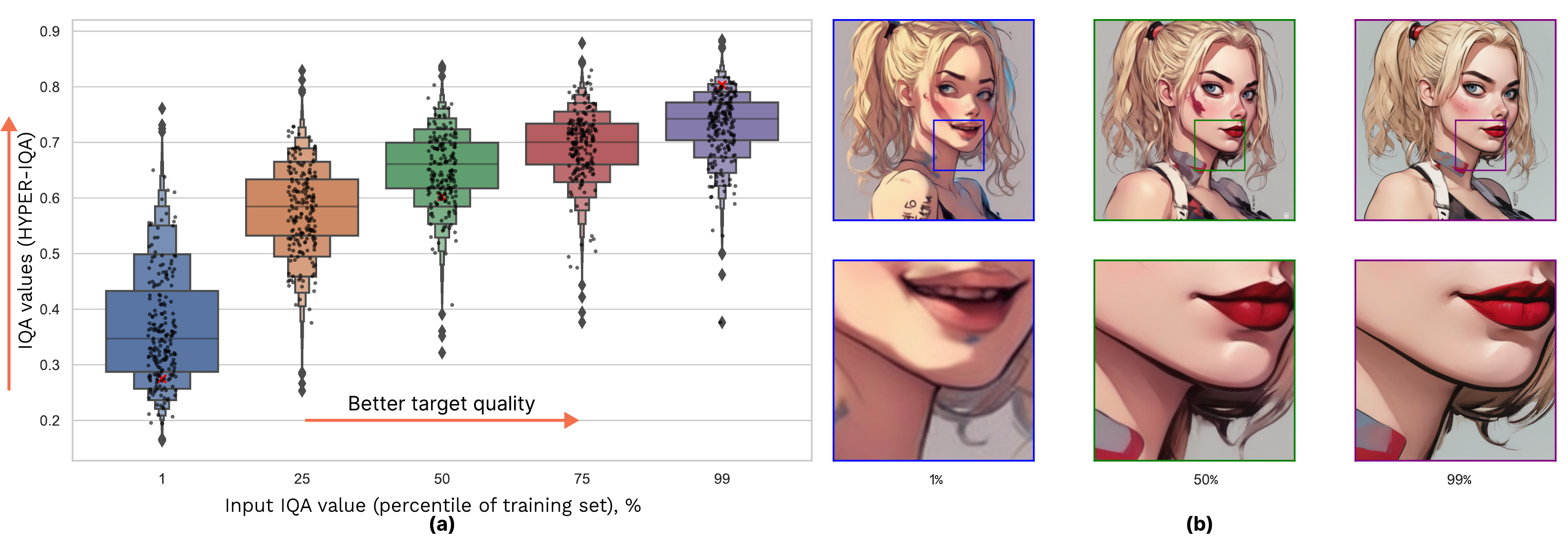}
     \caption{(a) Distributions of quality scores for images generated with the IQA-Adapter conditioned on different target quality levels (1-99 percentiles of the training dataset). The adapter is trained with the HYPER-IQA metric. (b) Examples of images generated with different target quality.}
  \label{fig:demo_quantiles}%
\vspace*{-0.4cm}
\end{figure*}


\textbf{IQA/IAA Models.} We experimented with a diverse set of 21 state-of-the-art quality assessment models, varying in architecture and training dataset. The models include CNN-based approaches like ARNIQA \cite{arniqa}, DBCNN \cite{zhang2020blind}, and CNNIQA \cite{kang2014convolutional}; TOPIQ \cite{topiq}, which combines a CNN backbone with an attention mechanism; HYPER-IQA \cite{Su_2020_CVPR}, which leverages a hyper-network with a CNN. Additionally, we tested transformer-based models, including MUSIQ \cite{ke2021musiq}, TRES \cite{golestaneh2021no}, and MANIQA \cite{yang2022maniqa} and metrics integrating vision-language capabilities like LIQE\cite{zhang2023liqe} and CLIP-IQA+\cite{clipiqa}. Where available, multiple versions of some models were tested, each trained on different datasets. Table \ref{tab:metrics_list} in the supplementary lists all used metrics with their corresponding training datasets.


\textbf{Evaluation Datasets.} We use several diverse prompt and image datasets for model evaluation:
\begin{itemize} 
\item \textit{Qualitative evaluation}: A filtered subset of 8,200 user-generated prompts from Lexica.art website \cite{lexica_dataset_hf} and PartiPrompts \cite{yu2022scaling} (1,600 prompts of different aspects and challenges).
\item \textit{Generative and compositional capabilities evaluation}: GenEval benchmark \cite{ghosh2024geneval} and corresponding prompts.
\item \textit{Additional fidelity measures}: A subset of 10,000 captions from  MS COCO \cite{lin2014microsoft} for FID \cite{heusel2017gans} and related scores.
\item \textit{Reference-based conditioning}: KADID-10k IQA dataset of 81 non-distorted images and 125 distorted variations for each source image (25 distortion types $\times$ 5 scales).
\end{itemize}

\subsection{High-quality conditioning}
\label{sec:hq_cond}
To evaluate the effectiveness of knowledge transfer from IQA models to diffusion-based generative models, we first explore the high-quality conditioning scenario, as this is the most intuitive application for quality-aware generation. To assess improvements objectively, we calculate the relative gain in quality scores compared to the base model:
\begin{equation*}
\operatorname{RelGain} = \frac{1}{N} \sum_{i=0}^N \frac{f(x_i') - f(x_i)}{f(x_i)} \cdot 100\%
\end{equation*}
where $f(x)$ denotes the quality assessment model, $x_i$ and $x_i'$ are images generated under the same prompt and seed for the base and quality-conditioned models, respectively. 



For IQA-Adapter, high-quality conditioning is achieved by setting the input to the 99-th percentile of the target metric’s values from the training dataset. Separate IQA-Adapters were trained for each IQA/IAA metric, and a multi-metric approach was tested by conditioning on combinations of different IQA/IAA models. 

Aside from the gradient-based baseline method of qualitative knowledge transfer, we compare IQA-Adapter with multiple existing methods of generation quality improvement of different nature: DiffusionDPO \cite{diffusion_dpo} (fine-tuning), Q-Refine \cite{q_refine} (Image Enhancement), BeautifulPrompt \cite{beautifulprompt} (prompt refactoring with LM) and simple textual tags emphasized with Prompt Weighting \cite{compel}. Figure \ref{fig:heatmap_adapter}(a) shows the relative gains for all evaluated methods on user-generated prompts from Lexica.art dataset (see Figure \ref{fig:heatmap_adapter_supp} in Supplementary for all tested IQA-Adapters). Detailed results for PartiPrompts dataset and the gradient-based method are provided in the Supplementary Section \ref{sec:hq_cond_supp}.

The gradient-based method, which directly optimizes IQA scores, increases target scores but generally fails to improve other IQA/IAA metrics, likely due to adversarial exploitation of model-specific vulnerabilities. Given its limitations, we focus on IQA-Adapter in the remaining experiments, discussing the use of the gradient-based method in adversarial scenarios in Section \ref{sec:disc_adv_patterns}. 

Unlike the gradient-based approach, the IQA-Adapters trained even on single IQA models show consistent quality gains across multiple metrics, with an average improvement of 7-9\% over the base model. Notably, gains for the target metric do not significantly exceed those for other metrics, demonstrating strong cross-metric transferability. 

Most IQA-Adapters demonstrate higher average quality gain compared to other existing methods, as well as the gradient-based approach. However, the best result is achieved by the combination of IQA-Adapter with DiffusionDPO and BeautifulPrompt, which signifies the mutual compatibility of the adapter with other approaches. IQA-Adapter trained with TOPIQ and LAION-AES metrics shows the most balanced results between technical quality and aesthetic scores, and qualitative negative guidance further improves average quality gain.

Notably, IQA scores tend to improve more easily than IAA scores, likely because IQA focuses on perceptual quality attributes that are less dependent on composition, whereas IAA is more content-sensitive and requires adjustments in both text and quality conditions.

Using multiple IQA/IAA metrics enhances the IQA-Adapter’s performance across evaluation metrics. For example, combinations like TOPIQ and LAION-AES models, and multiple versions of TOPIQ ("TOPIQ (4 versions)" row on Figure \ref{fig:heatmap_adapter}(a)), exhibit the best transferability, suggesting that diverse metrics provide richer quality information, broadening the IQA-Adapter’s capacity to capture complex qualitative attributes. Quality improvements of IQA-Adapter are further supported by a subjective study detailed in Section \ref{sec:qual_alignment}.


\subsection{Alignment with qualitative condition}
\label{sec:qual_alignment}
To assess the alignment between input quality conditions provided to the IQA-Adapter during generation and the quality of generated images, we attempt to condition it on different percentiles of the target IQA model’s values on the training dataset. Figure \ref{fig:demo_quantiles} demonstrates the impact of quality-condition on IQA scores and examples of images generated for corresponding quality levels. The results indicate a gradual increase in quality scores from the IQA model as the input condition rises, with generated images appearing progressively sharper and more detailed. We exemplify more quality conditions for the IQA-Adapters trained with different IQA/IAA models in supplementary Section \ref{sec:quality_guidance_examples_supp}.


\begin{table}[t]
\centering
\resizebox{1.0\columnwidth}{!}{%
\begin{tabular}{c|cccccc|c}
\toprule
\makecell{Models\\in IQA-Adapter} & \makecell{Two\\
Object}$\uparrow$ & \makecell{Attribute\\
Binding}$\uparrow$ & Colors$\uparrow$ & Counting$\uparrow$ & \makecell{Single\\
Object}$\uparrow$ & Position$\uparrow$ & Overall$\uparrow$ \\
\midrule
MANIQA (PIPAL) & 73.23\% & 20.25\% & 86.17\% & 36.56\% & 96.56\% & 10.50\% & 53.88\% \\
TOPIQ (KONIQ) & 71.97\% & 18.75\% & 85.11\% & 38.75\% & 98.12\% & \underline{13.75\%} & 54.41\% \\
CLIPIQA+, LIQE-MIX & 71.72\% & 20.25\% & 85.64\% & 41.25\% & 97.81\% & 11.75\% & 54.74\% \\
TOPIQ, LAION-AES & 69.70\% & 18.75\% & 85.90\% & \underline{45.31\%} & 99.38\% & 13.00\% & 55.34\% \\
\makecell{3xARNIQA,LIQE-MIX\\(different datasets)} & \textbf{73.99\%} & 19.25\% & \textbf{89.36\%} & 39.69\% & \textbf{99.69\%} & 13.75\% & 55.95\% \\
CLIP-IQA+ & 72.73\% & \underline{22.75\%} & 88.03\% & 43.44\% & 98.44\% & 12.25\% & 56.27\% \\
HYPER-IQA & \underline{73.99\%} & \textbf{25.25\%} & 85.90\% & 39.69\% & 98.75\% & \textbf{14.75\%} & \underline{56.39\%} \\
TOPIQ (FLIVE) & 72.73\% & 21.75\% & 87.77\% & \textbf{45.94\%} & 99.38\% & 13.00\% & \textbf{56.76\%} \\
\midrule
Base Model & 73.74\% & 21.75\% & \underline{88.30\%} & 43.75\% & \underline{99.69\%} & 10.50\% & 56.29\% \\
\bottomrule
\end{tabular}
}
\caption{Results of the IQA-Adapters trained with different IQA/IAA models on GenEval benchmark. Percents represent the accuracy of object-detection model on generated images. Results for all evaluated adapters are available in supplementary Table \ref{tab:geneval_supp}.}
\vspace*{-0.3cm}
\label{tab:geneval}
\end{table}

\textbf{Subjective Study}. To confirm that image quality improves with input quality conditions, we conducted a subjective study with the IQA-Adapter conditioned on three quality levels: low (1st percentile), medium (50th percentile), and high (99th percentile), as well as the base model (SDXL-Base). We utilized IQA-Adapter conditioned on TOPIQ and LAION-AES models, which showed the highest average IQA/IAA metric increases (Figure \ref{fig:heatmap_adapter}(a)). Participants evaluated the visual quality of images generated from 300 prompts, contributing over 22,300 responses from 1,017 users, with each image pair evaluated by at least 10 unique users (12.1 on average). For each model, we calculated the overall win rate defined as a share of image pairs on which it achieved the majority of votes. Additionally, we report the average percent of votes for the model across all image-pairs. Results are shown in Figure \ref{fig:heatmap_adapter}(b), and pairwise win rates in Figure \ref{fig:heatmap_adapter}(c). For more details on the subjective study, refer to Supplementary Section \ref{sec:subj_study_supp}.



Win rates align well with input quality conditions: high-quality conditions achieve the highest win rate, followed by medium- and low-quality. As shown in Figure \ref{fig:heatmap_adapter}(c), the IQA-Adapter conditioned on high quality outperforms the base model with 60\% win rate, compared to 32\% for the base model ($\sim$7\% were rated equally). This demonstrates that IQA-Adapter effectively captures and reproduces qualitative concepts aligned with human image quality judgments. Notably, the win rate for the low-quality condition drops significantly compared to medium quality. Figure \ref{fig:demo_quantiles}(a) further indicates that objective quality decreases sharply below the 25th percentile.

\subsection{Evaluating generative capabilities}
\label{sec:gen_cap}

To evaluate the generative capabilities of the quality-conditioned model and ensure that it doesn’t affect the ability to follow the textual prompt and generate diverse images, we tested it on the GenEval \cite{ghosh2024geneval} benchmark. It uses an object-detection model to evaluate the alignment between generated images and textual conditions. Table \ref{tab:geneval} shows the comparison results. Overall scores for most adapters are close to those of the base model. For each evaluation criterion, there is an IQA-Adapter that consistently outperforms the base model. The IQA-Adapter trained with HYPER-IQA, for example, increases “Attribute binding” (rendering two objects with two different colors) and “Position” (rendering two objects with specific relative positions) scores, suggesting better alignment with complex compositional prompts. The least improvement is in “Counting,” likely due to some IQA-Adapters’ tendency to add small details that sometimes increase object counts unnecessarily.

Additionally, we calculated FID, IS \cite{salimans2016improved} and CLIP \cite{hessel2021clipscore} scores for all tested adapters on a 10,000 captions subset of MS COCO. The results can be found in supplementary Section \ref{sec:eval_gen_cap_supp}. 
In summary, these findings indicate that the adapter conditioned on high quality mostly retains the generative capabilities of the base model, while shifting the generation towards a higher-quality subdomain.

\subsection{Reference-based qualitative conditioning}
\label{sec:ref_exps}
\begin{figure}
\hspace*{-0.5cm}
\begin{minipage}[b]{1.1\columnwidth}
\centering
   \includegraphics[width=0.99\linewidth]{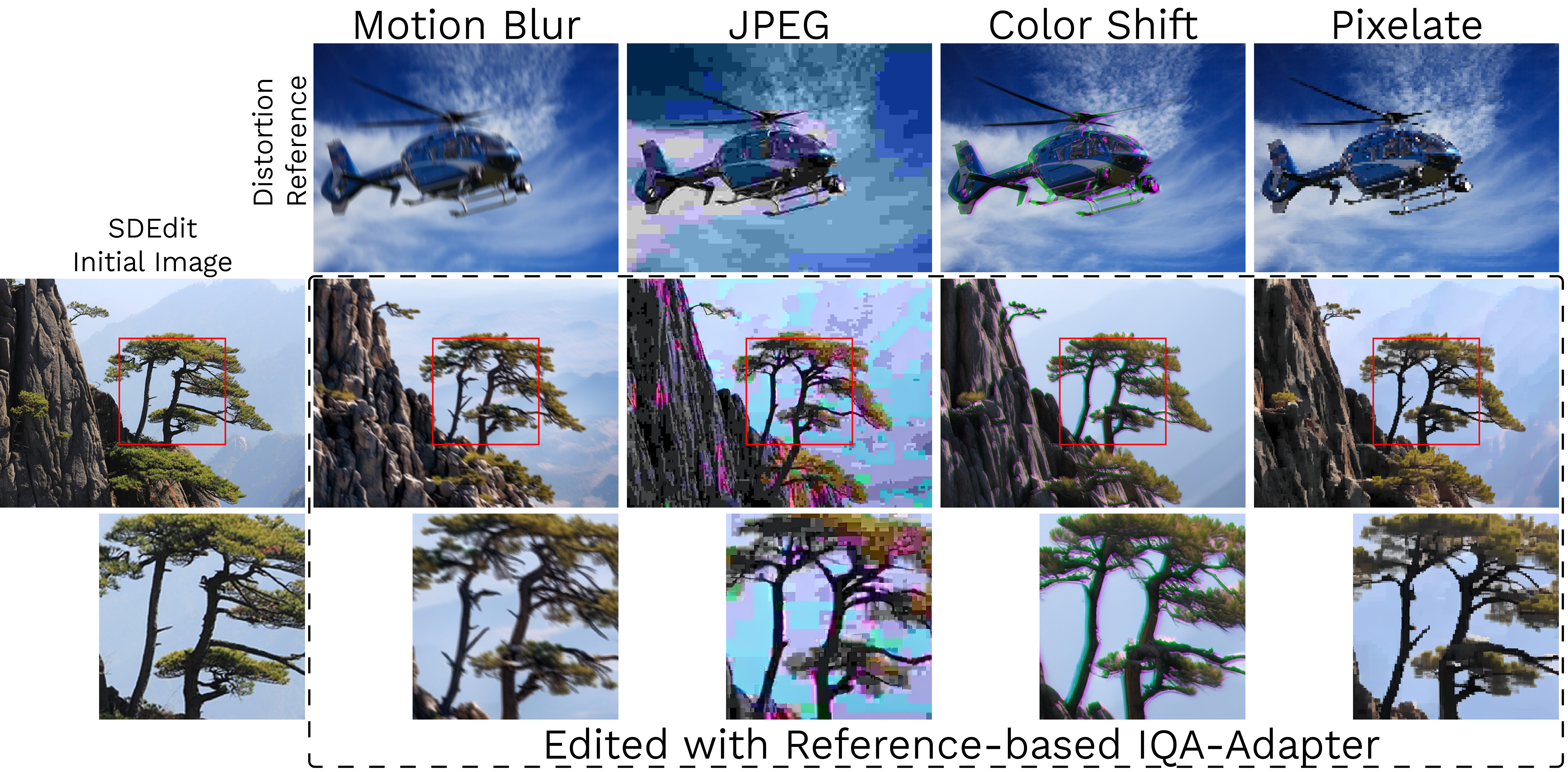}
  \captionof{figure}{Examples of Reference-based IQA-Adapter conditioning.}
  \label{fig:ref_iqa_adapter_example}%
\end{minipage}\hfill
\vspace*{-0.7cm}
\end{figure}

To test the ability of the Reference-based IQA-Adapter to transfer qualitative knowledge via IQA embeddings, we evaluate its ability to reproduce various image distortions present in the KADID-10k~\cite{kadid} dataset, which is often used to evaluate the performance of IQA metrics. The main goal of this experiment is to test if the adapter can distinguish the distortion on the reference image and transfer it to the generated one \textit{without} capturing any additional semantic information unrelated to image quality. Figure \ref{fig:ref_iqa_adapter_example} demonstrates Reference-based image editing using IQA-Adapter.

We select one of the 81 source images from KADID-10k and use all its distorted variations as references. These references guide the generation of images with corresponding distortion types and scales in two settings: Image-to-Image (I2I) editing using SDEdit~\cite{meng2022sdedit} and Text-to-Image (T2I) generation using synthetic captions from the BLIP-2~\cite{blip2} model. For I2I setting, the remaining 80 images serve as initializations for SDEdit, yielding 125 variations with different distortions per source image; and for T2I, BLIP-2 provides captions for the same 80 undistorted images, from which 125 variations per prompt are generated.

We compare Reference-based IQA-Adapter with IP-Adapter \cite{ye2023ip-adapter}, a common image-prompting technique, and StyleCrafter \cite{liu2023stylecrafter}, an adapter for artistic style transfer. Additionally, we evaluate IQA-Adapter that only accepts IQA scores of the reference image as a qualitative condition. To quantitatively evaluate the distortion transfer, we calculate multiple statistics: First, we measure CLIP-T and CLIP-I scores. CLIP-I scores are calculated both with a real image corresponding to the prompt and distortion (\textit{higher} score indicates better alignment with a source image we attempt to distort), and with the distortion reference (\textit{lower} score indicates less semantic information “leakage” from the reference). To evaluate the qualitative alignment between generated and reference images, we measure Spearman’s correlation coefficient between target IQA metric\footnote{For IP-Adapter and StyleCrafter, we use ARNIQA for SROCC calculation} values on generated images and distortion references. Additionally, we calculate cosine similarities between IQA model’s activations on these images. We use ARNIQA IQA model for embedding similarity in this experiment, as it shows good performance on KADID dataset and its activation space is optimized to differentiate different types of distortions \cite{arniqa}. 
\begin{table}[t]
\hspace*{-0.45cm}
\resizebox{1.1\columnwidth}{!}{%
\begin{tabular}{cc|ccccc}
\toprule
\makecell{Gen.\\method} & \makecell{Distortion\\transfer method} & \makecell{SROCC\\w/ dist. ref.} $\uparrow$ & \makecell{CLIP-T $\uparrow$ \\w/ caption} & \makecell{CLIP-I \\w/ real}$\uparrow$ & \makecell{CLIP-I $\downarrow$ \\w/ dist. ref.} & \makecell{IQA Embed.\\ Similarity} $\uparrow$ \\
\midrule
\multirow[c]{4}{*}{T2I} & IP-Adapter & 0.41 & 29.42 & 78.78 & 68.40 & 0.79 \\
& \makecell{StyleCrafter} & 0.53 & 31.81 & 83.74 & 60.51 & \underline{0.86} \\
 & \makecell{Ref.-based IQA-Adapter} & \textbf{0.80} & 32.21 & \textbf{85.75} & 58.62 & \textbf{0.91} \\
 & \makecell{IQA-Adapter\\(ARNIQA)}  & 0.53 & \textbf{32.35} & 84.12 & \textbf{56.92} & 0.70 \\
 & \makecell{IQA-Adapter\\(TOPIQ + LAION-AES)} & \underline{0.76} & \underline{32.29} & \underline{84.30} & \underline{57.27} & 0.72 \\
 \midrule
\multirow[c]{4}{*}{SDEdit I2I} & IP-Adapter & 0.24 & 31.66 & 90.17 & 60.20 & 0.73 \\
 & \makecell{Ref.-based IQA-Adapter} & \underline{0.69} & \textbf{31.91} & \textbf{91.26} & 58.87 & \textbf{0.86} \\
 & \makecell{IQA-Adapter\\(ARNIQA)}  & 0.17 & 31.84 & \underline{90.58} & \textbf{57.88} & 0.69 \\
 & \makecell{IQA-Adapter\\(TOPIQ + LAION-AES)} & \textbf{0.79} & 31.84 & 90.57 & \underline{57.95} & 0.73 \\
\bottomrule
\end{tabular}
}
\label{tab:kadid_exp}
\caption{Quantitative results of distortion transfer experiment on KADID-10k dataset. The best results are highlighted in \textbf{bold}, and second-best are \underline{underlined}.}
\vspace*{-0.3cm}
\end{table}

While IP-Adapter and StyleCrafter excel in their corresponding domains (general image prompting and style transfer accordingly), they are suboptimal for qualitative conditioning. They often fail to distinguish different distortions (e.g. blur and compression) and tend to copy objects and color schemes present on the distortion reference. Figures \ref{fig:i2i_extended} and \ref{fig:t2i_viz} in Supplementary compare the results of image editing and T2I generation with IQA-Adapter, IP-Adapter and StyleCrafter. Evaluation confirms this effect: IP-Adapter shows consistently higher CLIP similarity with a distortion reference image, indicating the replication of semantic information from the reference, and lower IQA embedding similarity and correlation, followed by StyleCrafter. On the other hand, Reference-based IQA-Adapter utilizes useful properties of the IQA embeddings and only transfers content-agnostic information. This ability to efficiently capture and simulate highly specific qualitative features can potentially be used as a data augmentation step for other I2I tasks.
\section{Conclusion}
In this work, we explored different techniques to transfer knowledge from image quality assessment models to diffusion-based image generators. 
We proposed a novel IQA-Adapter approach that allows the generator model to learn implicit connections between images and corresponding quality levels and enables quality-aware generation. 
Experiments and subjective evaluation showed that IQA-Adapter efficiently conditions the generation process in a way that aligns with human judgment, all while retaining the generative capabilities of the base model. Additionally, we demonstrate various applications of IQA-conditioned generation, including the improvement of quality of generated images and reference-based image degradation modeling. 
We further discuss the Future Work and use cases of IQA-conditioned generation in Section \ref{sec:discussion} and Limitations of the method in Section \ref{sec:limitations_supp}. Additionally, in Section \ref{sec:adv_patterns_supp} we investigate connections between quality-conditioning and adversarial robustness of IQA models.
{
    \small
    \bibliographystyle{unsrt}
    \bibliography{main}
}

\clearpage
\setcounter{page}{1}
\maketitlesupplementary
If you are viewing this document on Mac/iOS and have problems with Figure rendering, please either open the file "supplementary\_pdfa\_converted.pdf" instead of "supplementary.pdf" (however, due to the conversion, the hyperlinks in it do not function), or use a third-party pdf viewer.
\section{Contents}
Here we briefly summarize the contents of all sections in this supplementary file:
\begin{itemize}
 \item Section \ref{sec:discussion}: Discussion of the possible use-cases of IQA-Adapter and Future Work;
 \item Section \ref{sec:employed_metrics}: A detailed summary of all IQA/IAA models used in this study;
 \item Section \ref{sec:adapter_training_supp}: Details on IQA-Adapter training;
  \item Section \ref{sec:limitations_supp}: Limitations of IQA-Adapter;
 \item Section \ref{sec:abl_study}: Ablation Study on adapter design;
 \item Section \ref{sec:hq_cond_supp}: More results regarding high-quality conditioning experiments, visual comparison with other methods;
 \item Section \ref{sec:eval_gen_cap_supp}: Detailed results on generative capabilities of different methods;
 \item Section \ref{sec:alignment_supp}: Experiments regarding alignment with qualitative conditions;
 \item Section \ref{sec:deg_model}: Evaluation of image degradation with Full-Reference IQA metrics;
 \item Section \ref{sec:subj_study_supp}: More details on Subjective Study;
 \item Section \ref{sec:additional_exps}: Miscellaneous experiments: time measurements, generation consistency, examples of quality modulation;
 \item Section \ref{sec:adv_patterns_supp}: Some connections between quality optimisation and adversarial robustness;
 \item Section \ref{sec:ref_adapter_supp}: More examples of Reference-based IQA-Adapter and comparison with IP-Adapter and StyleCrafter.
\end{itemize}
\section{Discussion and Future Work}
\label{sec:discussion}
\subsection{IQA-Adapter as a degradation model}
\begin{figure*}[ht!]
\centering
   \includegraphics[width=.79\textwidth]{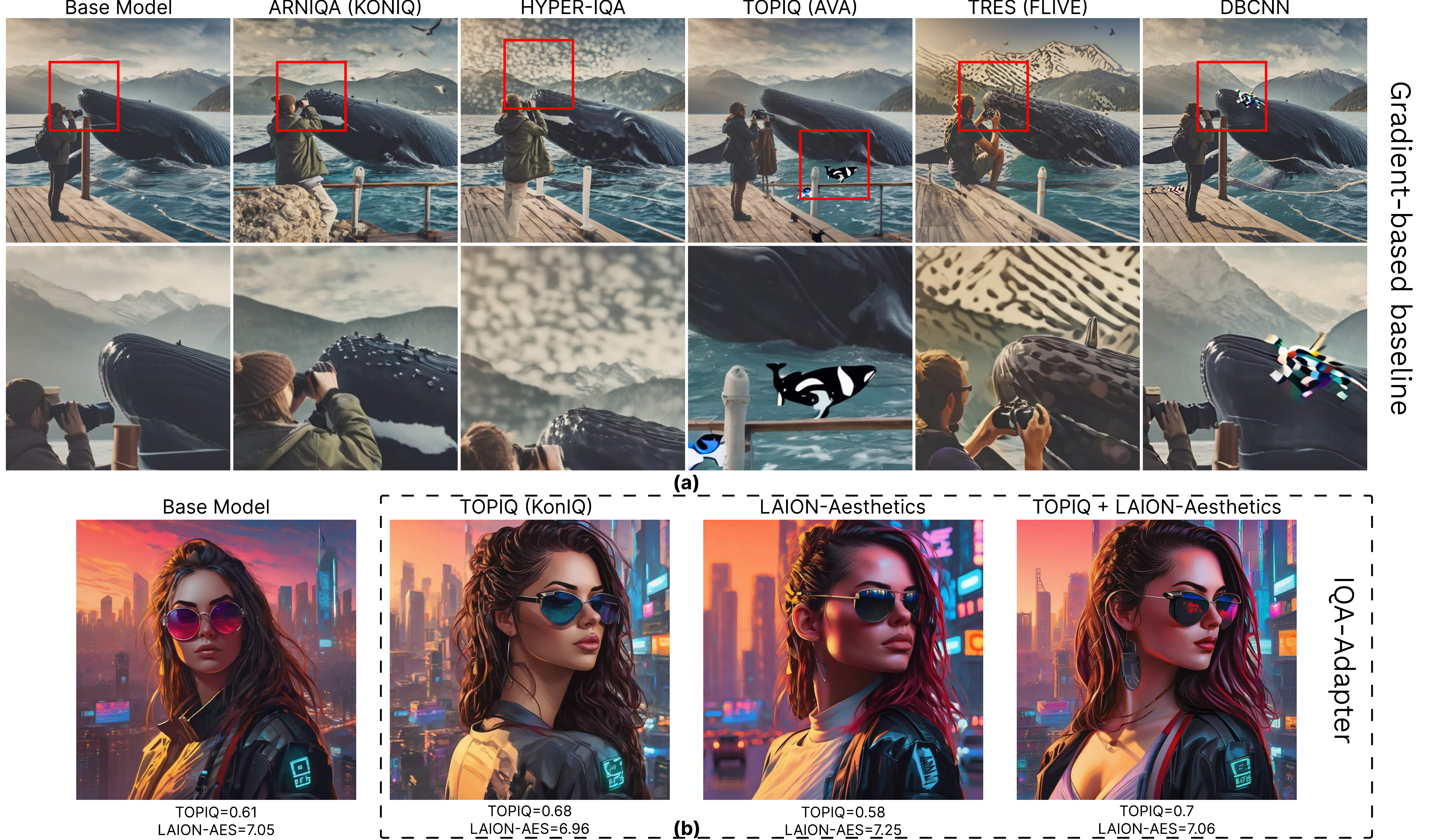}

  \caption{(a) Examples of adversarial patterns appearing under high \textbf{gradient-based} guidance scale. (b) Examples of images generated with the \textbf{IQA-Adapters} trained with different IQA models. Each IQA/IAA model has its stylistic preferences. All images in each line are generated with the same prompt and seed.}
  \label{fig:artifacts_prefs}%
\end{figure*}
\label{sec:degr_model}
As most IQA models are trained to assess distorted images, they can reliably detect noise, compression, blur, and other artifacts on images during IQA-Adapter training. Therefore, this knowledge is transferred to the generative model and such image attributes are connected with low-quality conditions. This allows IQA-Adapter to generate progressively more distorted images as input quality-condition decreases. The IQA-Adapter in Figure \ref{fig:demo_quantiles}(b), for example, implicitly learned to simulate JPEG compression artifacts when conditioned on low quality (1st percentile of the training dataset). Figure \ref{fig:loq_quality_viz} demonstrates more examples of similar artifacts appearing under low-quality guidance. As IQA models are mostly tailored to assess low-level quality attributes (in contrast with IAA methods), images produced with different quality levels usually retain similar content and composition, as illustrated in Figure \ref{fig:demo_2d} (bottom-to-top direction).

By applying appropriate filtering to exclude image pairs with unintended content differences, IQA-Adapter can generate large synthetic datasets of distorted and corresponding high-quality images. Such datasets can subsequently be used to pretrain models for image enhancement, deblurring, and other restoration tasks. While training such methods is a subject for future work, we additionally explore the distances between generated images with different target-quality conditions in Section \ref{sec:fr_metrics}. We also note that IQA-Adapter can be additionally fine-tuned with unpaired data containing specific distortions to simulate them during inference. 
\subsection{Exploring adversarial patterns and preferences of IQA models}
\label{sec:disc_adv_patterns}

When applied with a sufficiently high guidance scale, the gradient-based method can exploit vulnerabilities of the target IQA model, artificially inflating its values and shifting the generation towards an adversarial subdomain. This approach tends to produce images with distinct patterns specific to each IQA model. Figure \ref{fig:artifacts_prefs}(a) demonstrates adversarial patterns generated with different guidance models. For certain models, such as TRES and HYPER-IQA, these patterns form grid-like structures, and for others, like TOPIQ and DBCNN, they concentrate in smaller regions. We present more adversarial examples generated with gradient-based guidance and GradCAM \cite{jacobgilpytorchcam} visualizations of corresponding IQA models in Section \ref{sec:adv_patterns_supp}.

Our study further reveals that most IQA models exhibit distinct preferences when used with a high IQA-Adapter scale. For instance, TOPIQ often favors sharper images, while LAION-AES tends to enhance color saturation, producing more vibrant visuals. These effects can be compounded by using multiple IQA/IAA models simultaneously during adapter training, as illustrated in Figure \ref{fig:artifacts_prefs}(b). 



\section{Employed IQA/IAA methods}
\label{sec:employed_metrics}

Table~\ref{tab:metrics_list} provides a detailed summary of all IQA/IAA methods used in this study, along with their training datasets and architectural details. The column "PyIQA" lists model identifiers from the PyIQA library~\cite{pyiqa}. The column "Task" specifies supported tasks: most models are designed for IQA, while some (e.g., TOPIQ, MUSIQ) support both IQA and IAA, and others (e.g., NIMA) are exclusive to IAA. The column "Datasets" lists the datasets associated with each model; note that the models were not trained on mixtures of datasets, except for LIQE-MIX, which was specifically trained on a dataset mixture. For models like TOPIQ, there are several variants, each trained on a distinct dataset. The column "Arch" outlines the backbone architecture of the models. Most models are trained using finetuning of a pretrained model; however, some, like MUSIQ, are trained from scratch. The final three columns, "Params," "FLOPs," and "MACs," highlight the performance metrics of the models. FLOPs and MACs were computed using the calflops package~\cite{calflops}.

Table~\ref{tab:datasets_list} provides a detailed overview of the datasets used for training the IQA and IAA models. The column "Type" categorizes the datasets: FR indicates the presence of a distortion-free reference image used for collecting subjective scores, whereas NR denotes datasets without such references. The column "Year" indicates the release year of each dataset. The column "\# Ref" specifies the number of reference images used to generate distorted samples through augmentations. The column "\# Dist" represents the total number of samples in the dataset. The column "Dist Type." describes how distorted images were created: "synthetic" refers to distortions introduced via augmentations such as JPEG compression or blurring, "algorithmic" applies to distortions generated by neural networks, such as GAN-based modifications, "authentic" denotes images captured in natural, real-world conditions, and "aesthetics" refers to high-quality images sourced from stock photography collections. The column "\# Rating" indicates the number of ratings collected via crowdsourcing platforms. The column "Original size" details the resolution of images within the datasets.

\begin{table*}[hpt!]
\centering
\resizebox{0.99\textwidth}{!}{


\begin{tabular}{llcccrrr}
\toprule
\textbf{Model} 
    & \makecell{{PyIQA}} 
    & \makecell{{Task}} 
    & \makecell{{Datasets}} 
    & \makecell{{Arch}}        
    & {Params} 
    & \makecell{{FLOPS}} 
    & \makecell{{MACs}} \\
\midrule
TOPIQ~\cite{topiq}           
    & \texttt{topiq\_nr}      
    & IQA / IAA       
    & KonIQ-10k~\cite{hosu2020koniq}, SPAQ~\cite{spaq}, FLIVE~\cite{flive}, AVA~\cite{ava}                  
    & ResNet50             
    & 45.2M           
    & 886 GFLOPS     
    & 441.5 GMACs   \\
DBCNN~\cite{dbcnn}           
    & \texttt{dbcnn}          
    & IQA             
    & KonIQ-10k~\cite{hosu2020koniq}                                                  
    & VGG16                
    & 15.3M           
    & 2.1 TFLOPS     
    & 1 TMACs       \\
HyperIQA~\cite{hyperiqa}        
    & \texttt{hyper\_iqa}     
    & IQA             
    & KonIQ-10k~\cite{hosu2020koniq}                                                  
    & ResNet50             
    & 27.4M           
    & 2.6 TFLOPS     
    & 1.3 TMACs     \\
ARNIQA~\cite{arniqa}          
    & \texttt{arniqa}         
    & IQA             
    & KonIQ-10~\cite{hosu2020koniq}, FLIVE~\cite{flive},  KADID~\cite{kadid}                                  
    & ResNet50             
    & 23.5M           
    & \makecell{-}              
    & \makecell{-} \\
LIQE-Mix~\cite{zhang2023liqe}        
    & \texttt{liqe\_mix}      
    & IQA             
    & \makecell{Mixed \big(LIVE~\cite{live}, CSIQ~\cite{csiq}, KADID~\cite{kadid}, \\ CLIVE~\cite{clive},  BID~\cite{bid}, KonIQ-10k~\cite{hosu2020koniq} \big)}  
    & OpenAI CLIP ViT-B/32  
    & 151.3M          
    & 1.7 TFLOPS     
    & 850.7 GMACs   \\
MANIQA~\cite{yang2022maniqa}          
    & \texttt{maniqa}         
    & IQA             
    & KonIQ-10k~\cite{hosu2020koniq}, PIPAL~\cite{jinjin2020pipal}                                          
    & ViT-B/8              
    & 135.7M          
    & 56.4 TFLOPS    
    & 28.2 TMACs    \\
CNN-IQA~\cite{cnniqa}         
    & \texttt{cnniqa}         
    & IQA             
    & KonIQ-10k~\cite{hosu2020koniq}                                                  
    & CNN                  
    & 729.8K          
    & 49.4 GFLOPS    
    & 24.5 GMACs    \\
LIQE~\cite{zhang2023liqe}            
    & \texttt{liqe}           
    & IQA             
    & KonIQ-10k~\cite{hosu2020koniq}                                                  
    & OpenAI CLIP ViT-B/32 
    & 151.3M          
    & 1.7 TFLOPS     
    & 850.7 GMACs   \\
MUSIQ~\cite{musiq}          
    & \texttt{musiq}          
    & IQA / IAA       
    & KonIQ-10k~\cite{hosu2020koniq},  AVA~\cite{ava}, FLIVE~\cite{flive}                                  
    & Multiscale ViT       
    & 27.1M           
    & 400.6 GFLOPS   
    & 199.1 GMACs   \\
CLIP-IQA+~\cite{clipiqa}       
    & \texttt{cliq\_iqa+}     
    & IQA             
    & KonIQ-10k~\cite{hosu2020koniq}                                                  
    & OpenAI CLIP ResNet50 
    & 102.0M          
    & 981.1 GFLOPS   
    & 489.2 GMACs   \\
NIMA~\cite{nima}            
    & \texttt{nima}           
    & IAA             
    & AVA~\cite{ava}                                                        
    & InceptionResnetV2    
    & 54.3M           
    & 342.9 GFLOPS   
    & 171 GMACs     \\
LAION-Aes~\cite{laion_aes}       
    & \texttt{laion\_aes}     
    & IAA             
    & Other                                                      
    & OpenAI CLIP VIT-L/14  
    & 428.5M          
    & 2 TFLOPS       
    & 1 TMACs       \\
TReS~\cite{tres}            
    & \texttt{tres}           
    & IQA             
    & FLIVE~\cite{flive}                                                      
    & ResNet50             
    & 152.5M          
    & 25.9 TFLOPS    
    & 12.9 TMACs  \\
HPSv2~\cite{wu2023human}            
    & --          
    & Human Preference             
    & Human Preference Dataset v2~\cite{wu2023human}                                                      
    & OpenAI CLIP VIT-L/14  
    & 428.5M          
    & 2 TFLOPS       
    & 1 TMACs       \\\bottomrule
\end{tabular}

}
\caption{List of employed metrics with their corresponding training datasets.}
\label{tab:metrics_list}
\end{table*}

\begin{table*}[t]
\centering
\resizebox{0.8\textwidth}{!}{
    \begin{tabular}{clcccccc}
\toprule
Type 
    & Dataset 
    & Year
    &  \# Ref 
    & \# Dist 
    & Dist Type. 
    & \# Rating 
    & \makecell[c]{Original size \\ $W\times H$} \\
\midrule
    \multirow{4}{*}{FR} 
& LIVE~\cite{live}
    & 2006
    & 29 
    & 779 
    & Synthetic 
    & 25k 
    & $768\times512$ (typical) \\  
& CSIQ~\cite{csiq}
    & 2010
    & 30 
    & 866 
    & Synthetic 
    & 5k 
    & $512\times512$ \\ 
& KADID-10k~\cite{kadid}
    & 2019
    & 81 
    & 10.1k 
    & Synthetic 
    & 30.4k 
    & $512\times384$ \\ 
& PIPAL~\cite{jinjin2020pipal}
    & 2020
    & 250 
    & 29k 
    & Syth.+alg. 
    & 1.13M 
    & $288\times288$ \\ 
\midrule
\multirow{6}{*}{NR} 
& BID~\cite{bid}
    & 2010
    & 120
    & 6000 
    & Synthetic
    & $\sim$ 7k
    & 1K -- 2K \\
& AVA~\cite{ava}
    & 2012
    & -
    & 250k 
    & Aesthetic 
    & 53M 
    & $< 800$ \\
& CLIVE~\cite{clive}
    & 2015
    & -
    & 1.2k 
    & Authentic 
    & 350k 
    & $500\times500$ \\
& KonIQ-10k~\cite{hosu2020koniq}
    & 2018
    & -
    & 10k 
    & Authentic 
    & 1.2M 
    &  $512\times384$ \\
& SPAQ~\cite{spaq}
    & 2020
    & -
    & 11k 
    & Authentic 
    & -- 
    & 4K (typical) \\
& FLIVE~\cite{flive}
    & 2020
    & -
    & 160k 
    & Auth.+Aest. 
    & 3.9M 
    & Train$<640$ $\mid$ Test$>640$ \\
\bottomrule
\end{tabular}
}
\caption{Description of training datasets from Table~\ref{tab:metrics_list}.}
\label{tab:datasets_list}
\end{table*}


\section{IQA-Adapter training}
\label{sec:adapter_training_supp}
The IQA-Adapters were trained on the CC3M dataset, which consists of approximately 3 million text-image pairs, for 24,000 steps, followed by fine-tuning on a subset of the LAION-5B dataset, containing 170,000 images, for 3,000 steps. During training on CC3M, the images were center-cropped to a resolution of 512 $\times$ 512. For fine-tuning on LAION, the resolution was increased to 1024 $\times$ 1024 to match SDXL’s native resolution. We used the AdamW \cite{loshchilov2019decoupled} optimizer with $\beta_1=0.9$, $\beta_2=0.999$ and a weight decay of $1\times 10^{-2}$ for the IQA-Adapter parameters. All experiments utilized bf16 mixed precision to improve computational efficiency. Multi-node training was conducted using the accelerate \cite{accelerate} library, enabling efficient scaling across our hardware setup. We use batch\_size=16 per GPU for $512 \times 512$ training resolution, and batch\_size=4 for $1024 \times 1024$ fine-tuning. Each training run was launched on 5 nodes (40 GPUs). The learning rate was set to $10^{-4}$ during the primary training phase on CC3M and reduced to $10^{-5}$ for the fine-tuning on the LAION subset. For Reference-based IQA-Adapter, we apply series of degradations to training images with a probability $p=0.1$ during training.

To ensure consistency and reproducibility, all experiments were conducted within Docker containers built from a shared image. The environment included Python 3.11, PyTorch 2.1, and other dependencies required for training and inference. We use adapter scale $\lambda=0.5$ in all experiments, unless stated otherwise, and negative guidance scale $
\delta=0.3$, if IQA-Adapter name includes "+ Neg. G." ($\delta=0$ otherwise). For Reference-based IQA-Adapter, we use adapter scale $\lambda=0.65$.

\section{Limitations}
\label{sec:limitations_supp}
IQA-Adapter serves as a guiding mechanism for transferring knowledge from the IQA/IAA domain to generative models. However, the extent of this knowledge transfer is inherently constrained by the capabilities and limitations of current IQA/IAA models. Most existing IQA datasets, and the models trained on them, are designed to assess the quality of real images, focusing on aesthetical attributes and distortions common for human-generated images. These models often lack the ability to detect distortions specific to generated content, such as unnatural or anatomically incorrect features (e.g., distorted limbs or physically implausible scenes). As a result, these issues may not be adequately penalized in the quality estimates used for guidance, limiting the adapter’s ability to address such generation defects. One possible direction of future work to address this limitation is to train a classifier for different kinds of generation artifacts and then attempt to utilize its logits as a conditioning factor.

Another limitation arises from biases in the training data. The IQA-Adapter can inadvertently learn and reproduce unintended relationships between image content and quality levels present in the dataset. For example, when conditioned on low aesthetic scores, the adapter may occasionally generate images with watermarks, likely because it encountered numerous stock photos with watermarks during training and associated them with lower-quality conditions. While some of these correlations may be considered genuine (e.g., watermarks generally reduce image aesthetics), such artifacts highlight the challenge of disentangling genuine quality attributes from dataset-specific correlations.

The training process itself introduces additional challenges. IQA-Adapter training occurs entirely in the latent space of the diffusion model, while the quality scores used for supervision are computed in pixel space. This discrepancy between the latent representations of images (compressed by the model's VAE encoder) and the pixel-level quality scores can introduce instability into the training process, as the adapter must work with imperfect representations of the input images. Furthermore, the VAE decoder used in the final generation step imposes inherent limitations, as it may introduce artifacts (e.g., blurred text or texture inconsistencies) that the adapter cannot correct. In this work, we only cover existing quality assessment models; however, this limitation can be largely mitigated in the future by implementing a quality assessment model that operates in the latent space of the generative model.
\begin{figure*}[ht!]
\centering
   \includegraphics[width=.85\textwidth]{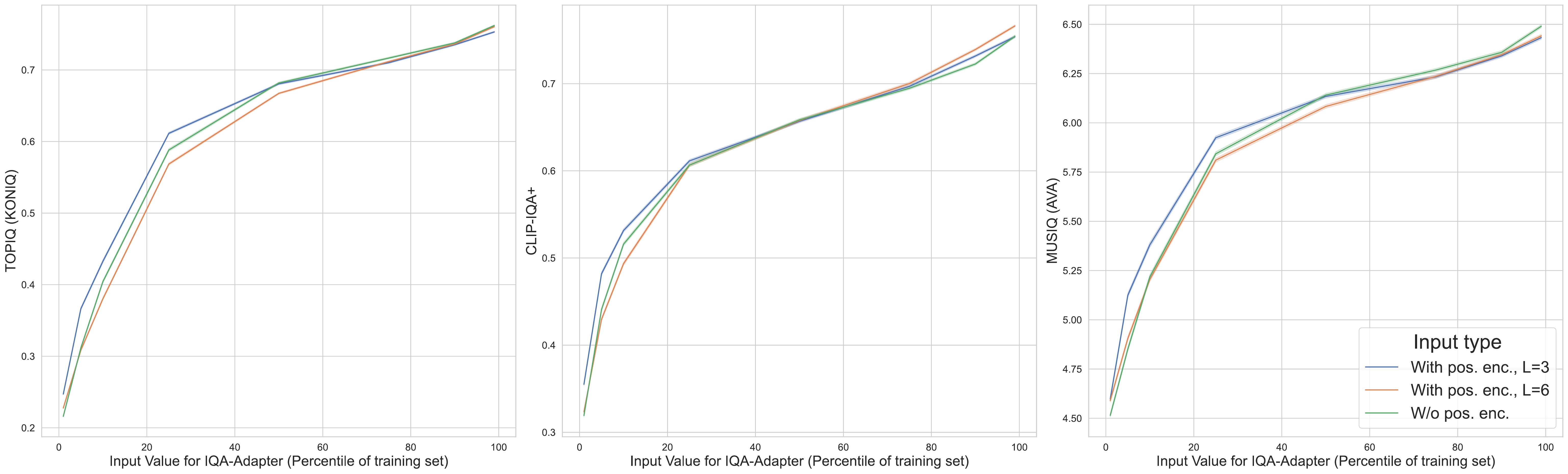}
  \caption{Results of the IQA-Adapter modulation on input quality-condition for different types of input preprocessing with positional encoding. For all evaluated types, adapter was trained with TOPIQ (KonIQ) model.}
  \label{fig:posenc_exp}%
\end{figure*}
\section{Ablation Study}
\label{sec:abl_study}
In this section, we report the results of our experiments with different architectural elements and hyperparameters of the IQA-Adapter. We compare our base design with a "simplified" model (Sec. \ref{sec:abl_study_dca}) and a more sophisticated approach with Positional Encoding (Sec. \ref{sec:pos_enc}). Furthermore, we evaluate the impact of the scaling hyperparameter $\lambda$ of IQA-Adapter. 

\subsection{Impact of the Separate Qualitative Attention and Negative Guidance}
\label{sec:abl_study_dca}
\begin{table}[ht!]
\centering
\resizebox{1.0\columnwidth}{!}{%
\begin{tabular}{c|cc|ccccc}
\toprule
Model & \makecell{Quality \\Gain, \%}$\uparrow$ & \makecell{SROCC \\w/ target}$\uparrow$ & FID$\downarrow$ & \makecell{FID\\(TOP-10\%)}$\downarrow$ & IS $\uparrow$ & CLIP-T $\uparrow$ & CLIP-I $\uparrow$ \\
\midrule
IQA-Adapter & 8.95 & 0.97 & \textbf{21.36} & \textbf{28.44} & \textbf{36.89} & \textbf{26.83} & \textbf{70.02} \\
\makecell{IQA-Adapter\\+ Neg. Guidance} & \textbf{10.86} & \textbf{0.98} & 22.16 & 29.25 & 36.33 & 26.80 & 69.82 \\
\makecell{IQA-Adapter \\w/o Separate Cross-Attn} & 8.31 & 0.26 & 29.04 & 39.91 & 30.22 & 26.34 & 67.9 \\
\bottomrule
\end{tabular}
}
\caption{Comparison of IQA-Adapters with and without separate qualitative attention. Both adapters are trained with TOPIQ and LAION-Aesthtics IQA models. SROCC is calculated with target TOPIQ scores, and Quality Gain is evaluated similarly to Sec. \ref{sec:hq_cond} and averaged across all evaluation metrics.}
\label{tab:ablation_dca}
\end{table}

To test the importance of the separate qualitative cross-attention operation, we test the ablated IQA-Adapter that simply concatenates qualitative tokens to the text ones and processes them within a single (textual) cross-attention operation. This simplified model functionally resembles “adaptive” Textual Inversion \cite{galimage}, controlled by a projection module.

In this setting, adapter loses the ability to control its impact via $\lambda$ parameter, reducing its usability. As demonstrated in Table \ref{tab:ablation_dca}, the model partially retains the ability for qualitative improvements; however, qualitative prompt-following capabilities of the simplified model greatly diminish, as evidenced by reduced correlation between target and predicted quality of the generated images: it drops from 0.97 to 0.27 SROCC. Furthermore, simultaneous processing of the new tokens with contextual information reduces the textual prompt-following capabilities of the model, as evidenced by FID and CLIP scores. This emphasizes the importance of the attention separation for qualitative conditioning. It also demonstrates that the the disengagement of qualitative and contextual information is beneficial for learning content-independent relationships between quality-related image properties.

\subsection{Positional Encoding}
\label{sec:pos_enc}

Given that the quality metrics used as input for the IQA-Adapter form a low-dimensional representation (e.g., a 2D space for quality and aesthetics, as shown in Figure \ref{fig:demo_2d}), we explored the use of positional encoding to enrich these inputs. Inspired by the sinusoidal encoding strategy employed in NeRFs\cite{nerf} and timestamp encoding in Stable Diffusion models\cite{rombach2022high}, we applied the following transformation to each input IQA/IAA value independently:
\begin{equation*}
\begin{aligned}
\gamma (x) = \big(x, \sin(2^0 \pi x), \cos(2^0 \pi x), \ldots, \\
\sin(2^{L-1} \pi x), \cos(2^{L-1} \pi x)\big),
\end{aligned}
\end{equation*}
where \( x \) is the input value, and \( L \) controls the number of additional components in the representation. All IQA/IAA inputs were normalized to zero mean and unit variance prior to this transformation.  

We hypothesized that positional encoding would enhance the model’s sensitivity to subtle quality variations, allowing for more fine-grained control over output quality without affecting behavior at the edges of the input range. However, our experiments demonstrated that positional encoding had minimal impact on the model's behavior.  

To evaluate this, we conducted experiments where the IQA-Adapter was modulated on the input quality condition, as described in Sections \ref{sec:qual_alignment} and \ref{sec:alignment_supp}. Using a dataset of user-generated prompts from Lexica.art, we compared IQA-Adapters with and without positional encoding across a range of evaluation metrics. The results, shown in Figure \ref{fig:posenc_exp}, indicate that positional encoding produced outcomes nearly identical to those of the baseline IQA-Adapter, regardless of the value of \( L \).  

Although our experiments did not reveal significant benefits from positional encoding for the quality-conditioning task, we believe there may be potential for improvement with alternative encoding strategies. For instance, rotary positional embeddings (RoPE)\cite{rope}, which have shown success in recent large language models, could be a promising direction. We leave the exploration of such strategies for future research.

\subsection{Impact of IQA-Adapter scaling factor}
\label{sec:scale_exp}
\begin{figure*}[ht!]
\centering
   \includegraphics[width=.98\textwidth]{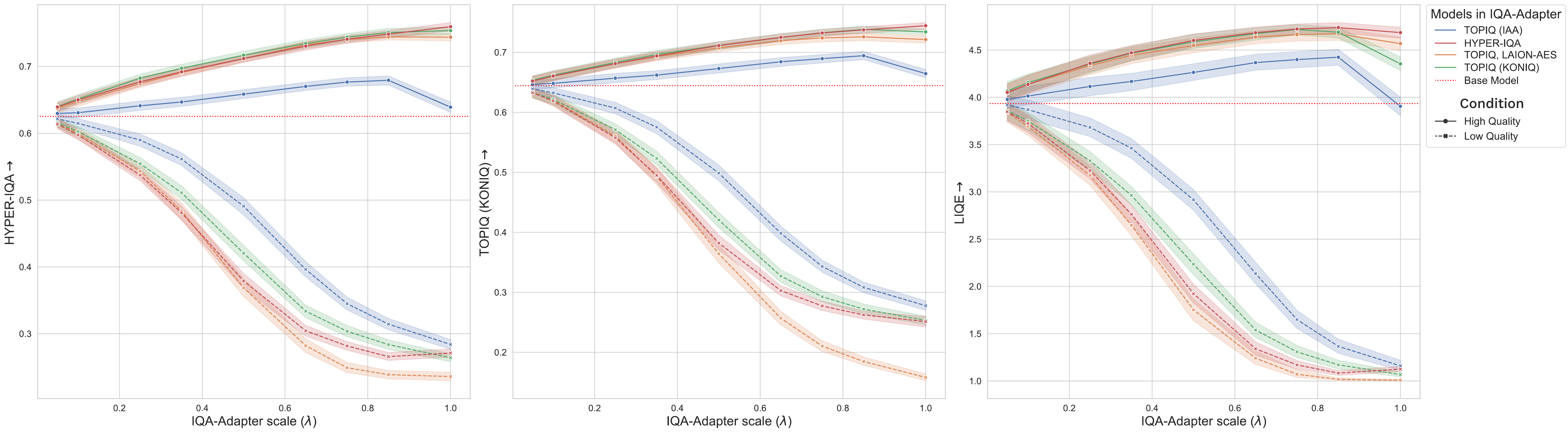}
  \caption{The relationship between image-quality scores (evaluated by the HYPER-IQA, TOPIQ and LIQE metrics) and the adapter scale parameter ($\lambda$) for the IQA-Adapters trained with different target IQA/IAA models and conditioned on low (dashed line) and high (solid line) target quality. For reference, the red dotted line indicates the quality level of the base model. The experiment utilized 300 random user-generated prompts from the Lexica.art dataset.}
  \label{fig:scale_exp_lineplots}%
\end{figure*}

To evaluate the impact of the adapter scale parameter \( \lambda \) on the visual quality of generated images, we tested IQA-Adapters trained with various IQA/IAA models under both high- and low-quality input conditions. We evaluated 9 $\lambda$ values ranging from 0.05 to 1.0. For each configuration, images were generated using 300 randomly sampled prompts from the Lexica.art dataset. The results are shown in Figure \ref{fig:scale_exp_lineplots}.

As \( \lambda \) increases, image quality scores deviate progressively from the base model's levels, aligning with the specified quality condition. Under high-quality conditions, the increase in quality is smooth and resembles a logarithmic curve for most adapters, reflecting diminishing returns as the base model already achieves relatively high-quality outputs. Beyond a certain threshold for \( \lambda \), typically around 0.75, further increases cease to improve quality, with excessively high values (\( \lambda > 0.9 \)) introducing artifacts that reduce both visual quality and IQA/IAA scores.

In low-quality conditions, the quality degradation progresses more rapidly, as the adapter has greater freedom to modify the image. The decrease in scores follows a sigmoidal trend: minimal change occurs for small \( \lambda \) values, but the effect accelerates significantly beyond \( \lambda \sim 0.4 \) and plateaus at the adapter's limits near \( \lambda \sim 0.75-0.85 \). This behavior highlights the non-linear relationship between adapter strength and its impact on image quality, with optimal performance generally observed for \( \lambda \) values in the range of [0.5, 0.75] for both low- and high-quality conditioning.

\section{High-quality conditioning: more results}
\label{sec:hq_cond_supp}
\subsection{Gradient-based guidance}
\label{sec:hq_gen_guidance_supp}
Figure \ref{fig:heatmaps_partiprompts}(b) presents the relative gain in metric scores when using the gradient-based approach to optimize image quality during generation for prompts from PartiPrompts \cite{yu2022scaling}. Unlike IQA-Adapter, direct optimization of the target metric improves that specific metric alone, while most other quality metrics tend to decline. This observation highlights the adversarial nature of gradient-based guidance, further confirmed by a closer examination of changes in generated images, which reveal adversarial patterns (as shown in Figure \ref{fig:adv_examples}). Interestingly, certain metrics, such as ARNIQA (trained on KADID), LAION-AES, and LIQE MIX, show improvements even when unrelated quality metrics are targeted for optimization. This behavior points to their inherent instability and susceptibility to adversarial attacks, raising questions about their robustness as quality measures.

\subsection{IQA-Adapter}
\label{sec:hq_gen_adapter_supp}
\begin{figure*}[ht!]
\centering
   \includegraphics[width=.98\textwidth]{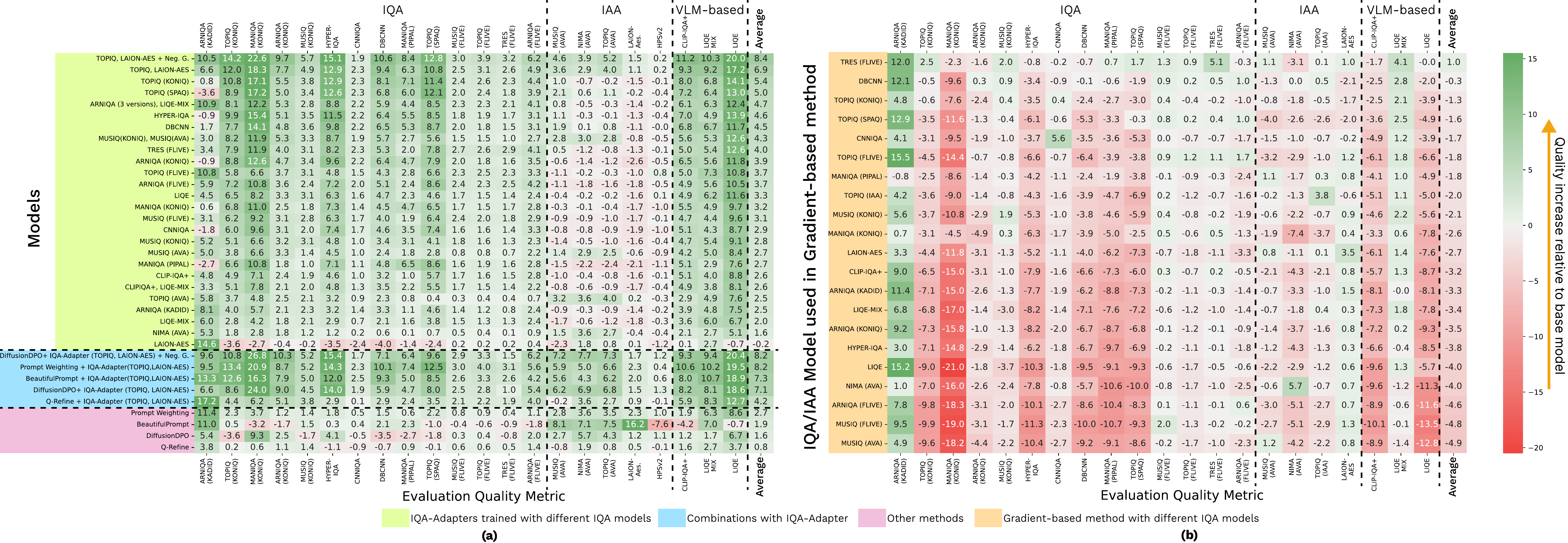}
  \caption{Quality improvement relative to base model (in \%) for the IQA-Adapters trained on different IQA/IAA
models and other generation quality improvement methods (a); and gradient-based method targeted on different IQA/IAA models (b). All IQA-Adapters are conditioned with high target quality (99th percentile of the training dataset) and use the same prompts and seeds. Prompts are taken from PartiPrompts dataset.}
  \label{fig:heatmaps_partiprompts}%
\end{figure*}

\begin{figure*}[ht!]
\centering
   \includegraphics[width=.7\textwidth]{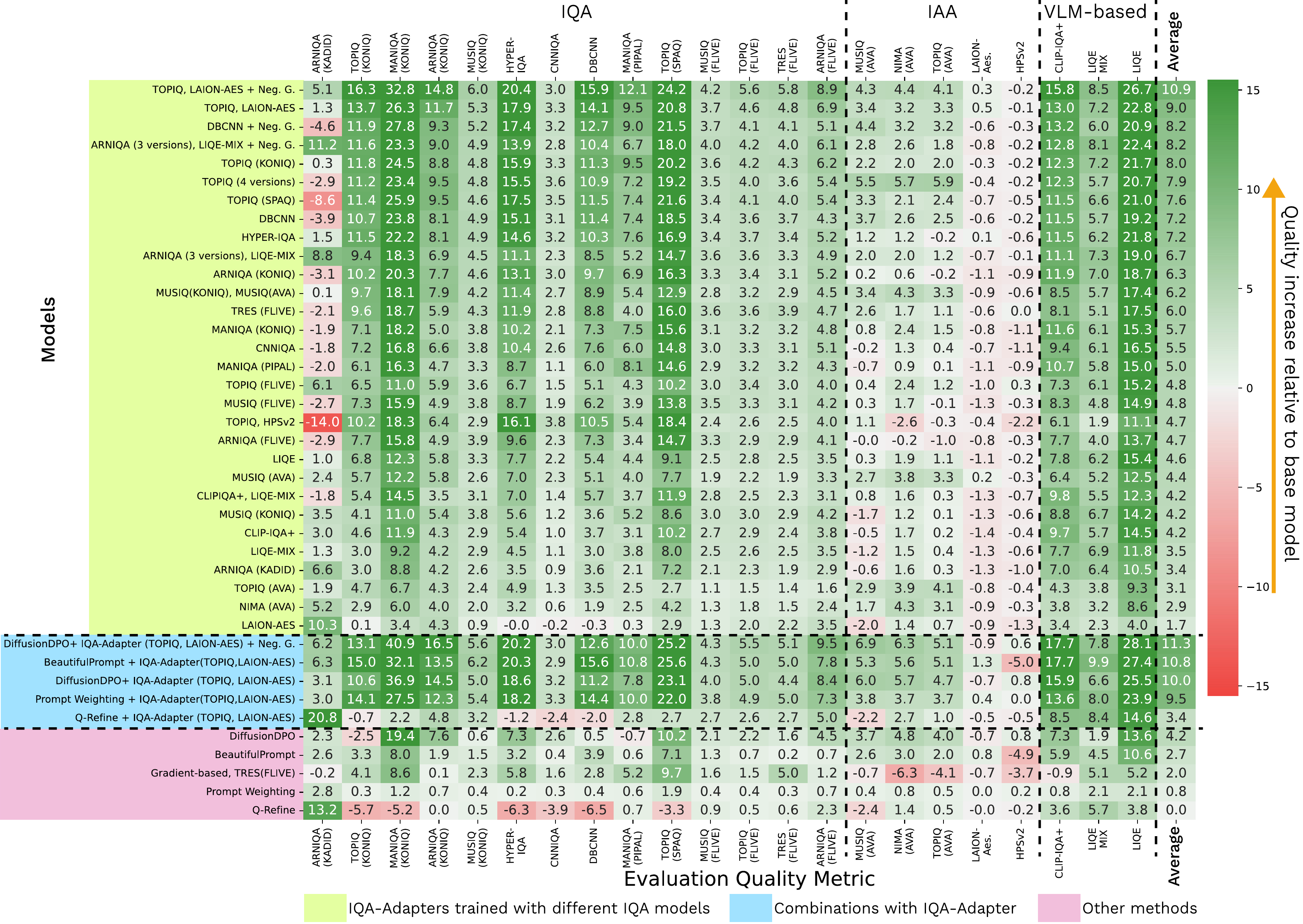}
  \caption{Quality improvement relative to base model (in \%) for the IQA-Adapters trained on different IQA/IAA
models and other generation quality improvement methods on Lexica.art dataset. This Figure complements the results reported in Figure \ref{fig:heatmap_adapter} in the main paper.}
  \label{fig:heatmap_adapter_supp}%
\end{figure*}

Figure \ref{fig:heatmap_adapter_supp} presents detailed results for all tested IQA-Adapters on Lexica.art dataset, complementing Figure \ref{fig:heatmap_adapter} (a) from the main paper. Figure \ref{fig:heatmaps_partiprompts} (a) provides additional results of high-quality conditioning with IQA-Adapter on PartiPrompts. The results on this dataset mirror the trends observed on the Lexica.art prompts, discussed in Section \ref{sec:hq_cond}. Specifically, conditioning on the 99th percentile of target metrics not only boosts the target metrics themselves but also improves most other metrics, highlighting the strong transferability of IQA-Adapter. However, the average metric improvements on PartiPrompts are 1–2\% lower than those observed on Lexica.art. This discrepancy can likely be attributed to the quality and completeness of the prompts. Unlike the more detailed and descriptive prompts in Lexica.art, PartiPrompts consists of shorter and more generic prompts. These simpler prompts impose fewer demands on the generation process, limiting the need for detailed generation, which is one of a key factors behind the significant metric improvements achieved by IQA-Adapter on Lexica.art.

Figure \ref{fig:comparison} demonstrates the comparison of IQA-Adapter with existing generation quality improvement methods on prompts sampled from Lexica.art dataset. IQA-Adapter conditioned on high quality usually results in more sharper and detailed results.

\section{Evaluating Generative Capabilities: more results}
\label{sec:eval_gen_cap_supp}
\begin{figure*}[ht!]
\centering
   \includegraphics[width=.98\textwidth]{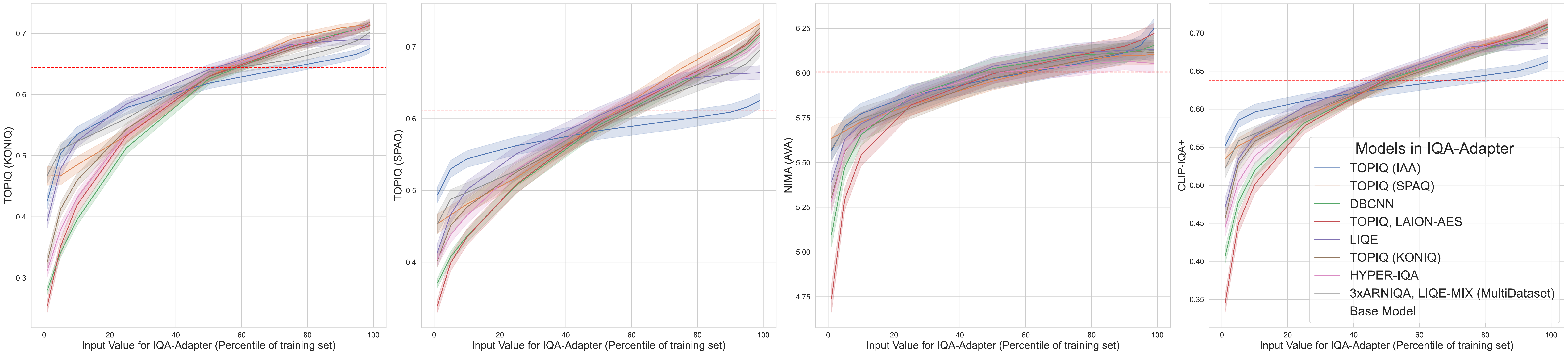}
  \caption{The relationship between input quality-condition (represented as a percentile of target IQA/IAA model on the training dataset) and image-quality scores evaluated by four different metrics (TOPIQ (KonIQ), TOPIQ (SPAQ), CLIP-IQA+, LIQE).}
  \label{fig:lineplots_quantiles}%
\end{figure*}

Table~\ref{tab:geneval_supp} provides the complete results on the GenEval benchmark. Among the 25 evaluated IQA-Adapters, five outperform the Base Model in terms of the overall score. Notably, even the weakest IQA-Adapter surpasses the Base Model in the Counting and Position metrics. However, the best-performing IQA-Adapter underperforms the Base Model in the Two Object, Colors, and Single Object metrics. Overall, while all IQA-Adapters achieve performance levels comparable to the initial model, some manage to outperform it in specific areas.

Table~\ref{tab:fid_is_clip} presents quantitative results for the FID, IS, and CLIP-similarity metrics. With a few exceptions, most IQA-Adapters exhibit slightly higher FID scores on the full MS COCO training dataset compared to the Base Model. This can be attributed to the diverse quality distribution of the dataset, which contains images of varying visual fidelity. Since IQA-Adapters are conditioned to prioritize high-quality generation, they naturally shift the output distribution toward a more specific subdomain characterized by higher visual quality. As a result, the distance to the broader, more heterogeneous image distribution of the full dataset increases. To address this domain shift, we also calculate FID scores on high-quality subsets of the MS COCO training dataset. These subsets include the top 10\% and 25\% of images, selected based on average quality scores from multiple IQA and IAA models. In this scenario, most IQA-Adapters consistently achieve lower FID scores than the Base Model, demonstrating superior alignment with the high-quality subsets.

In addition to FID, we evaluate the Inception Score (IS) and CLIP-similarity metrics. CLIP-Text (CLIP-T) measures the similarity between generated images and their corresponding text prompts, using COCO captions as prompts in our experiment. CLIP-Image (CLIP-I) measures the distance between generated images and the real images corresponding to the captions. Results indicate that most IQA-Adapters achieve better CLIP scores than the Base Model, highlighting improved prompt-following capabilities. However, the Inception Score results are slightly lower compared to the Base Model. It is worth noting that the IS differences fall within the confidence interval. Additionally, IS is not well-suited for evaluating SDXL model, which is trained on large-scale internet datasets~\cite{betzalel2022study}. Furthermore, as IQA-Adapters generate more complex and detailed images, the classifier behind Inception Score struggles to identify the main object within the scene, further complicating its evaluation.


\begin{table*}[t]
\centering
\resizebox{0.9\textwidth}{!}{

\begin{tabular}{llllllll}
\toprule
\makecell{Models\\in IQA-Adapter} 
    & \makecell{Two\\
Object}$\uparrow$ 
    & \makecell{Attribute\\
Binding}$\uparrow$ 
    & Colors$\uparrow$ 
    & Counting$\uparrow$ 
    & \makecell{Single\\
Object}$\uparrow$ 
    & Position$\uparrow$ 
    & Overall$\uparrow$ \\
\midrule
LAION-AES & 65.40\% & 16.75\% & 84.57\% & 45.00\% & 97.50\% & 12.25\% & 53.58\% \\
MANIQA (PIPAL) & 73.23\% & 20.25\% & 86.17\% & 36.56\% & 96.56\% & 10.50\% & 53.88\% \\
ARNIQA (FLIVE) & 69.70\% & 18.50\% & 84.04\% & 42.50\% & 97.81\% & 12.25\% & 54.13\% \\
TOPIQ (KONIQ) & 71.97\% & 18.75\% & 85.11\% & 38.75\% & 98.12\% & 13.75\% & 54.41\% \\
CLIPIQA+, LIQE-MIX & 71.72\% & 20.25\% & 85.64\% & 41.25\% & 97.81\% & 11.75\% & 54.74\% \\
LIQE-MIX & 68.43\% & 19.50\% & 87.50\% & 43.12\% & 98.12\% & 12.75\% & 54.91\% \\
MUSIQ (FLIVE) & 69.19\% & 23.25\% & \underline{88.30\%} & 39.38\% & 99.06\% & 12.50\% & 55.28\% \\
TOPIQ (4 versions) & 72.47\% & 21.75\% & 87.77\% & 40.31\% & 97.19\% & 12.25\% & 55.29\% \\
TOPIQ, LAION-AES & 69.70\% & 18.75\% & 85.90\% & 45.31\% & 99.38\% & 13.00\% & 55.34\% \\
TOPIQ(KONIQ), HPSv2 & 71.21\% & 22.25\% & 85.64\% & 42.50\% & 98.44\% & 12.25\% & 55.38\% \\
CNNIQA & 71.72\% & 19.50\% & 87.50\% & 41.56\% & 98.12\% & \underline{14.25\%} & 55.44\% \\
MUSIQ (AVA) & 69.44\% & 24.25\% & 86.97\% & 40.94\% & 99.06\% & 12.50\% & 55.53\% \\
MUSIQ(KONIQ), MUSIQ(AVA) & 73.23\% & 22.75\% & 86.44\% & 40.94\% & 98.12\% & 12.50\% & 55.66\% \\
TOPIQ (SPAQ) & 73.48\% & 21.25\% & 86.70\% & 43.75\% & 97.50\% & 12.50\% & 55.86\% \\
ARNIQA (3 versions), LIQE-MIX & 73.99\% & 19.25\% & \textbf{89.36\%} & 39.69\% & \textbf{99.69\%} & 13.75\% & 55.95\% \\
MANIQA (KONIQ) & 73.48\% & 25.75\% & 88.30\% & 38.75\% & 96.88\% & 12.75\% & 55.98\% \\
LIQE & 72.73\% & 21.75\% & 86.97\% & 41.56\% & 98.75\% & \underline{14.25\%} & 56.00\% \\
NIMA (AVA) & 70.96\% & 23.00\% & 87.50\% & 44.69\% & 98.44\% & 11.50\% & 56.01\% \\
MUSIQ (KONIQ) & 73.74\% & 21.00\% & 86.44\% & \underline{46.25\%} & 97.50\% & 11.50\% & 56.07\% \\
ARNIQA (KONIQ) & 71.97\% & 22.00\% & 87.50\% & 44.38\% & 98.12\% & 12.75\% & 56.12\% \\
CLIP-IQA+ & 72.73\% & 22.75\% & 88.03\% & 43.44\% & 98.44\% & 12.25\% & 56.27\% \\
HYPER-IQA & 73.99\% & 25.25\% & 85.90\% & 39.69\% & 98.75\% & \textbf{14.75\%} & 56.39\% \\
DBCNN & 73.48\% & 22.75\% & 86.44\% & 44.38\% & 99.06\% & 13.00\% & 56.52\% \\
ARNIQA (KADID) & 72.98\% & 23.25\% & 86.97\% & 45.94\% & 98.75\% & 11.50\% & 56.56\% \\
TOPIQ (AVA) & 75.00\% & 22.50\% & 87.77\% & 42.81\% & 98.12\% & 13.50\% & 56.62\% \\
TOPIQ (FLIVE) & 72.73\% & 21.75\% & 87.77\% & 45.94\% & 99.38\% & 13.00\% & 56.76\% \\
\midrule
Base Model & 73.74\% & 21.75\% & 88.30\% & 43.75\% & \underline{99.69\%} & 10.50\% & 56.29\% \\
\midrule
DiffusionDPO & \textbf{83.33\%} & \underline{26.50\%} & 87.77\% & \underline{47.81\%} & \underline{99.69\%} & 12.50\% & \underline{59.60\%} \\
Q-Refine & 70.96\% & 21.75\% & \underline{88.83\%} & 40.94\% & 99.06\% & 9.75\% & 55.21\% \\
Prompt Weighting & 71.21\% & 23.00\% & 87.23\% & 43.12\% & 99.38\% & 11.50\% & 55.91\% \\
BeautifulPrompt & 18.94\% & 1.00\% & 35.90\% & 9.38\% & 72.81\% & 4.75\% & 23.80\% \\
\midrule
DiffusionDPO + IQA-Adapter (TOPIQ, LAION-AES) & \underline{83.08\%} & \underline{26.50\%} & 87.77\% & 45.94\% & 99.06\% & 13.75\% & \underline{59.35\%} \\
DiffusionDPO + IQA-Adapter(TOPIQ, HPSv2) & \underline{80.30\%} & \textbf{31.00\%} & 86.97\% & \textbf{50.62\%} & 99.06\% & 12.50\% & \textbf{60.08\%} \\
Q-Refine + IQA-Adapter (TOPIQ, LAION-AES) & 68.94\% & 19.00\% & 86.70\% & 44.69\% & 98.44\% & 11.75\% & 54.92\% \\

\bottomrule
\end{tabular}

}
\caption{GenEval, more results. The best results are \textbf{bold}, the second- and third-best are \underline{underlined}. Table is sorted over "Overall" column.}
\label{tab:geneval_supp}
\end{table*}

\begin{table*}[t]
\centering
\resizebox{0.9\textwidth}{!}{

\begin{tabular}{c|ccc|ccc}
\toprule
Models in IQA-Adapter 
    & \makecell{FID$\downarrow$\\Full} 
    & \makecell{FID$\downarrow$\\(Top-25\%)} 
    & \makecell{FID$\downarrow$\\(Top-10\%)} 
    & IS$\uparrow$ 
    & CLIP-T$\uparrow$ 
    & CLIP-I$\uparrow$ \\
\midrule
LAION-AES & 23.94 & 28.96 & 34.53 & 34.27±0.85 & 26.73 & 69.75 \\
MUSIQ(KONIQ), MUSIQ(AVA) & 22.48 & 24.96 & 29.68 & 37.00±1.43 & 26.79 & 69.47 \\
NIMA (AVA) & 22.32 & 25.65 & 30.55 & 37.72±1.08 & 26.70 & 69.80 \\
TRES (FLIVE) & 22.27 & 22.82 & 27.21 & 37.90±0.76 & 26.50 & 69.52 \\
TOPIQ (AVA) & 22.25 & 25.50 & 30.40 & 36.86±0.94 & 26.78 & 69.83 \\
ARNIQA (3 versions), LIQE-MIX & 21.95 & 22.92 & 27.58 & 37.55±1.02 & 26.69 & 69.62 \\
TOPIQ (4 versions) & 21.93 & 23.69 & 28.32 & 36.99±1.76 & 26.79 & 69.74 \\
MANIQA (KONIQ) & 21.74 & 23.85 & 28.57 & 37.63±1.23 & 26.91 & 69.61 \\
CLIPIQA+, LIQE-MIX & 21.43 & 22.45 & 27.02 & 38.33±1.83 & 26.70 & 69.65 \\
TOPIQ, LAION-AES & 21.36 & 23.53 & 28.44 & 36.89±1.33 & 26.83 & \underline{70.02} \\
MUSIQ (AVA) & 21.20 & 24.92 & 30.08 & 36.42±1.39 & 26.93 & 69.96 \\
ARNIQA (KONIQ) & 21.13 & 22.70 & 27.53 & 37.32±0.87 & 26.86 & 69.53 \\
TOPIQ (FLIVE) & 21.04 & \textbf{21.63} & \textbf{26.28} & 37.93±0.70 & 26.64 & 69.54 \\
HYPER-IQA & 21.00 & 22.82 & 27.69 & 37.99±1.19 & 26.90 & 69.26 \\
DBCNN & 20.85 & 22.43 & 27.20 & 38.28±1.44 & 26.84 & 69.60 \\
MUSIQ (KONIQ) & 20.77 & 22.38 & 27.08 & \underline{38.57±1.12} & 26.80 & 69.55 \\
LIQE & 20.76 & 22.34 & 27.21 & 37.72±1.46 & 26.82 & 69.81 \\
CLIP-IQA+ & 20.45 & 21.89 & \underline{26.55} & 37.66±1.05 & 26.80 & \textbf{70.05} \\
ARNIQA (FLIVE) & 20.44 & \underline{21.75} & \underline{26.58} & 38.25±1.20 & 26.85 & \underline{69.99} \\
ARNIQA (KADID) & 20.35 & 22.50 & 27.56 & 37.67±1.31 & 26.76 & 69.32 \\
LIQE-MIX & 20.35 & 22.26 & 27.18 & 38.09±1.02 & 26.79 & 69.65 \\
TOPIQ (SPAQ) & 20.28 & 22.85 & 27.79 & 37.07±1.12 & 26.84 & 69.26 \\
TOPIQ (KONIQ) & 20.17 & 21.95 & 26.90 & 37.29±1.15 & \underline{26.96} & 69.49 \\
TOPIQ, HPSv2 & \underline{19.67} & 22.08 & 27.40 & 36.71±1.45 & \underline{27.00} & 69.12 \\
CNNIQA & \underline{19.61} & 22.40 & 27.53 & 37.87±1.18 & 26.94 & 69.31 \\
MANIQA (PIPAL) & \textbf{19.27} & \underline{21.88} & 27.19 & 37.98±1.46 & 26.77 & 69.48 \\
\midrule
Base Model & 19.92 & 23.15 & 28.41 & \textbf{39.44±1.66} & 26.70 & 69.35 \\
\midrule
BeautifulPrompt & 30.92 & 35.64 & 40.83 & 33.30±1.12 & 21.23 & 58.01 \\
DiffusionDPO & 29.57 & 34.04 & 38.88 & 36.93±1.04 & \textbf{27.10} & 68.74 \\
Prompt Weighting & 24.14 & 26.02 & 30.50 & 38.44±2.15 & 26.42 & 68.78 \\
Q-Refine & 20.29 & 23.41 & 28.56 & \underline{39.05±1.11} & 26.83 & 69.11 \\
\bottomrule
\end{tabular}

}
\caption{FID, IS and CLIP scores of the IQA-Adapters trained with different IQA/IAA models on 10k subset of the MS COCO captions. FID-Full is calculated with the full MS COCO training dataset, and FID Top-n\% measures FID to the highest-quality subset of MS COCO (as measured by the average score across all IQA/IAA metrics) of the corresponding size. The best results are \textbf{bold}, the second- and third-best are \underline{underlined}. Table is sorted over "FID Full" column.}
\label{tab:fid_is_clip}
\end{table*}

\section{Alignment with qualitative condition: more results}
\label{sec:alignment_supp}
\begin{figure*}[ht!]
\centering
   \includegraphics[width=.65\textwidth]{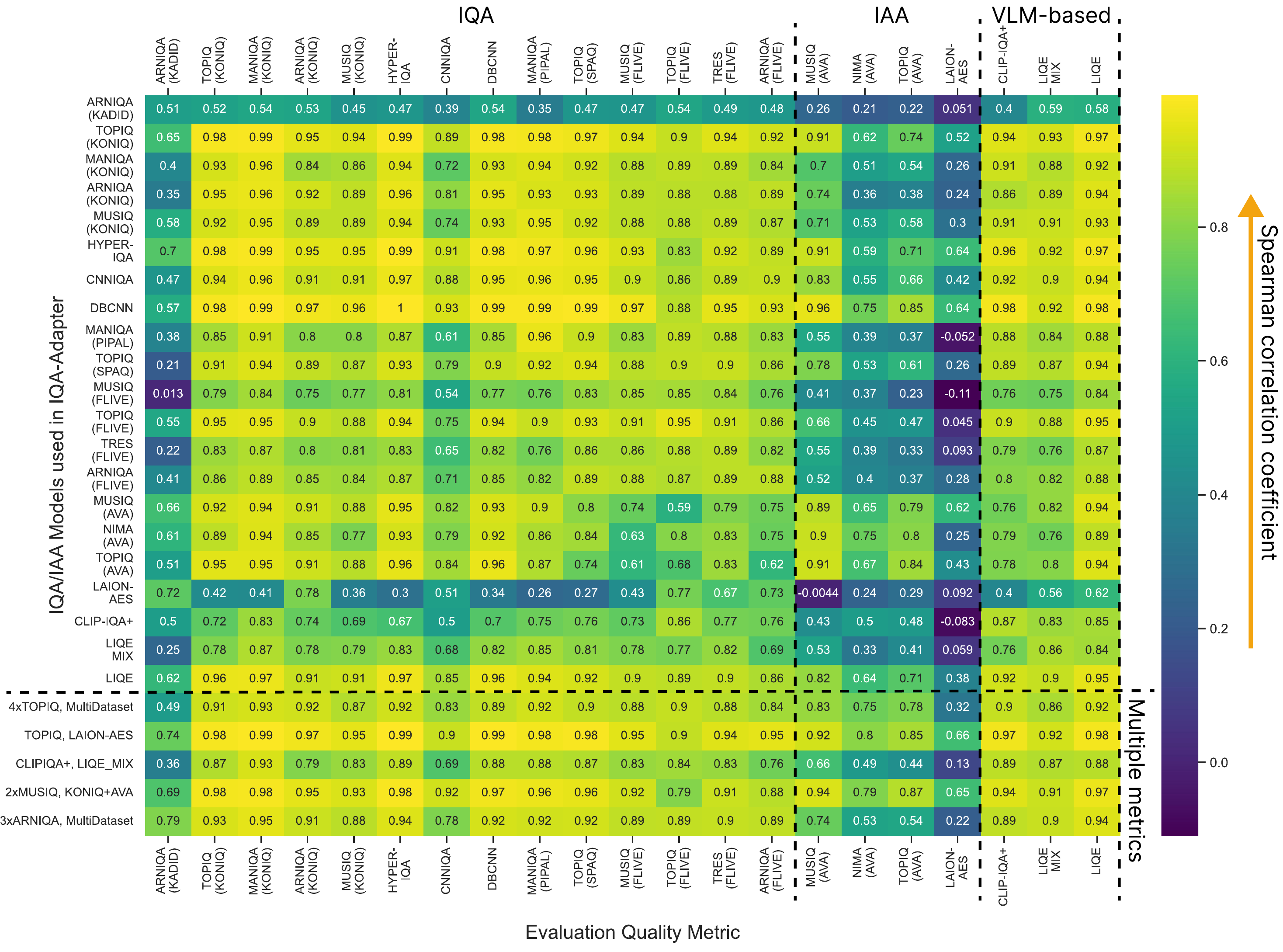}
  \caption{Correlations between input quality-conditions (represented as a percentile of target IQA/IAA model on the training dataset) and metric scores for the IQA-Adapters trained with different IQA/IAA models. Rows represent various IQA-Adapters, and columns indicate an IQA/IAA model used for SROCC calculation.}
  \label{fig:corrs_heatmap}%
\end{figure*}

To further evaluate the relationship between the input quality conditions provided to the IQA-Adapter during image generation and the quality of the resulting images, we analyzed correlations between the target quality and various metric scores. Figure \ref{fig:corrs_heatmap} shows estimated correlations for each trained IQA-Adapter. Generally, the metrics demonstrate a strong alignment with the target quality, with the highest correlations observed when comparing different IQA models. In contrast, weaker correlations are noted when IQA models are compared with IAA models. Among the evaluated metrics, the poorest correlations are associated with images generated using the IQA-Adapter based on the IAA metric, LAION-Aes. Interestingly, even the metric's own values fail to exhibit significant correlation, which may be attributed to the IQA-Adapter training process, specifically the additional fine-tuning step. However, when LAION-Aes is paired with an IQA metric, the correlations with IAA models improves significantly. For example, the IQA-Adapter trained on the TOPIQ and LAION-Aes metrics achieves high correlations with both IQA and IAA models, making it an optimal choice for generating images with high visual quality.

Additionally, Figure \ref{fig:lineplots_quantiles} illustrates the relationship between the average scores of four metrics and the input-quality conditions across different IQA-Adapters. All metrics show a monotonic increase in their mean scores, reinforcing the strong correlations shown in Figure \ref{fig:corrs_heatmap}. This trend is consistent across all IQA-Adapter types, regardless of whether they are trained on IQA models, IAA models, or VLM-based approaches. Starting from a specific target percentile --- typically around the 50th percentile --- the mean metric scores surpass those of the base model.

\section{IQA-Adapter as a degradation model}
\label{sec:deg_model}
\subsection{Examples of progressive quality degradation}
\label{sec:exampled_deg}
Figures \ref{fig:viz_quantiles} and \ref{fig:viz_quantiles_new} illustrate the generation results for different percentiles of metric scores on the training dataset. As the percentile decreases, the generated images begin to exhibit various distortions, such as compression artifacts, noise, blurring, and others. These distortions are likely present in the corresponding training datasets for the metrics, causing them to become sensitive to these distortions and assign lower scores. By passing progressively lower scores to the adapter, we can approximate a continuous path in the image-space between low and high-quality images on the ends of the spectrum. This qualitatively monotonic "path" (albeit with occasional local content changes) can potentially be used to train iterative image refinement algorithms.

\begin{figure*}[ht!]
\centering
   \includegraphics[width=.98\textwidth]{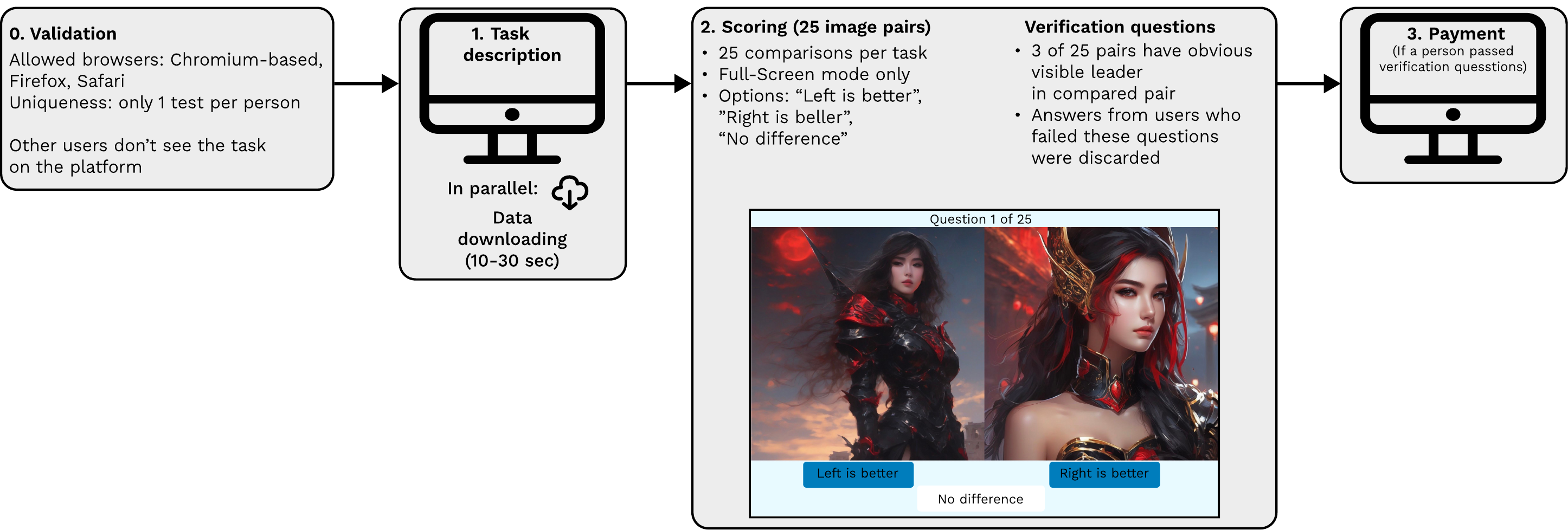}
  \caption{Overall scheme of the subjective study described in Sections \ref{sec:qual_alignment} and \ref{sec:subj_study_supp}.}
  \label{fig:subj_scheme}%
\end{figure*}

This quality-modulation ability of IQA-Adapter enables leveraging diffusion models as degradation models to generate various distortions, including natural ones. To achieve this, the IQA-Adapter should be trained on a dataset containing the relevant distortions, using as guidance either subjective assessments or a specialized metric sensitive to these distortions. Exploring this approach will be the focus of our future research. 

\begin{figure*}[ht!]
\centering
   \includegraphics[width=.85\textwidth]{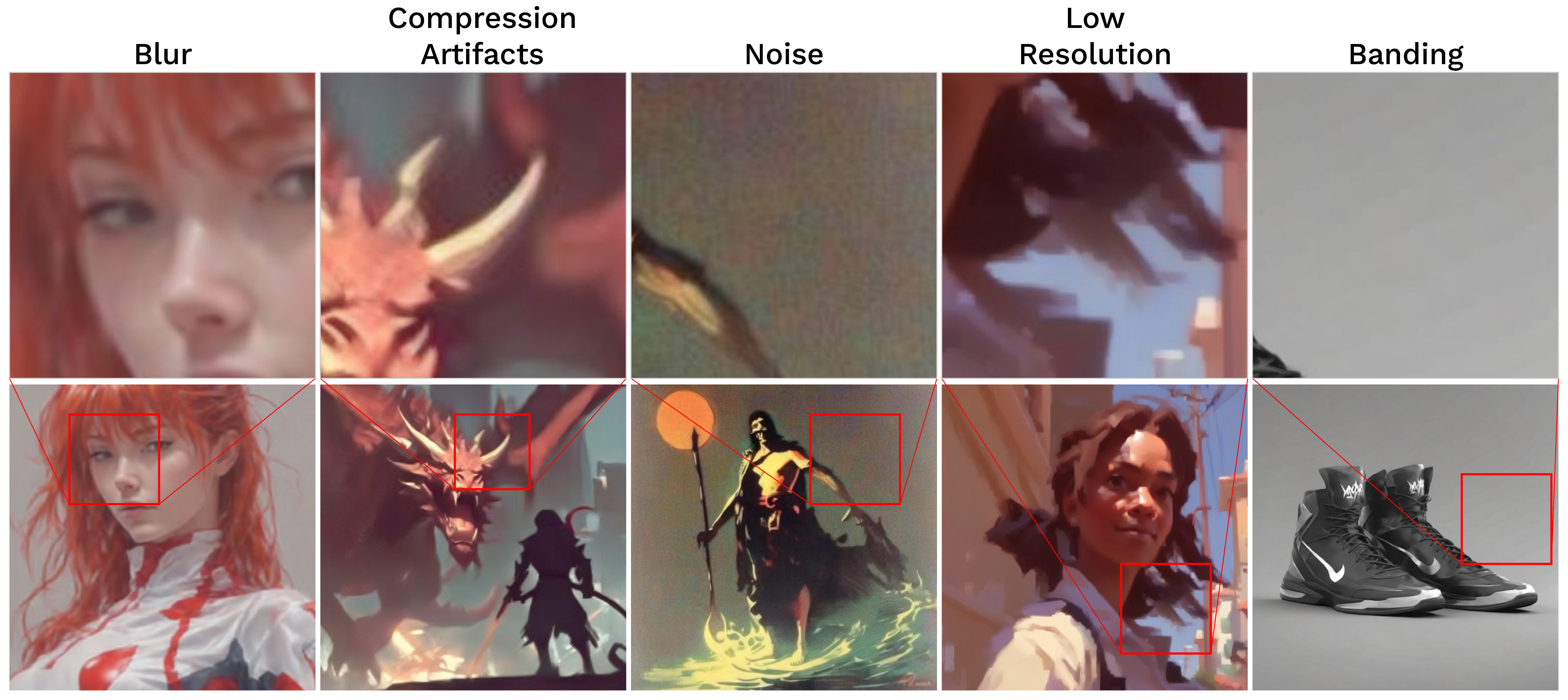}
  \caption{Examples of images generated with IQA-Adapter conditioned on \textbf{low} quality. IQA-Adapter is able to reproduce various distortions present in the training dataset.}
  \label{fig:loq_quality_viz}%
\end{figure*}

Figure \ref{fig:loq_quality_viz} presents additional examples of generated distortions under low-quality conditioning. Furthermore, section \ref{sec:ref_adapter_supp} provides visualizations of Reference-based IQA-Adapter conditioning on different specific distortions.
\subsection{Evaluating distances between high- and low-quality-conditioned generation}
To investigate the differences between images generated with varying target quality levels, we estimated the distances between them using four FR IQA metrics: SSIM \cite{ssim}, LPIPS \cite{lpips}, DISTS \cite{ding2020iqa}, and PieAPP \cite{Prashnani_2018_CVPR}. SSIM is a classical nonparametric method based on scene statistics, designed to assess structural similarity. LPIPS, on the other hand, is a neural network-based metric that measures similarity as the cosine distance between the features extracted from a pre-trained convolutional network. DISTS refines LPIPS by incorporating additional insensitivity to small image shifts, making it more robust. Lastly, PieAPP demonstrates strong correlations with subjective scores, particularly for the super-resolution (SR) task \cite{msu-superres}.

\begin{figure*}[ht!]
\centering
   \includegraphics[width=.98\textwidth]{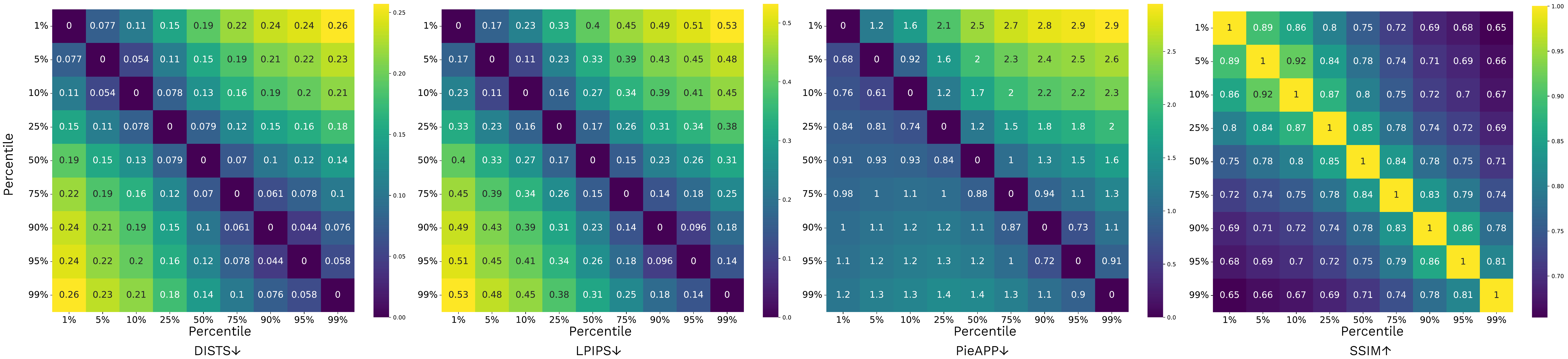}
  \caption{FR IQA metrics distances between images generated with the IQA-Adapter conditioned on different target-quality levels. The IQA-Adapter is trained for HYPER-IQA model.}
  \label{fig:frs_fig}%
\vspace{-0.4cm}
\end{figure*}

We generated 8,200 images with user-generated prompts from the Lexica.art website for each target quality level (percentile of metric scores on the training dataset). Figure \ref{fig:frs_fig} shows the average distances between corresponding images across different percentiles, measured using the selected FR metrics. As the gap between percentiles increases, the distance between them grows consistently as well. High-quality percentiles (90, 95, 99) are the closest to each other, whereas distant percentiles (e.g., 1 and 99) differ significantly, mostly because of the introduced semantic variations. In contrast, the nearest 2–3 percentiles are quite similar, with differences primarily in small details. Notably, DISTS shows lower differences than LPIPS, suggesting the presence of minor content shifts between images in different percentiles.

\label{sec:fr_metrics}


\section{Subjective Study}
\label{sec:subj_study_supp}

Our subjective study employed 300 randomly sampled user-generated prompts from the Lexica.art dataset. We used Subjectify.us platform for the evaluation. Overall scheme of the subjective study and the example of the user interface is demonstrated on Figure \ref{fig:subj_scheme}. During this study, we collected more than 22,300 valid responses of 1,017 unique users: each image-pair was independently assessed by at least 10 unique participants. As we compared 4 models (3 quality-conditions for the IQA-Adapter and the base model), total number of compared image-pairs was $\frac{4\cdot3}{2} \times 300=1800$. Participants were asked to evaluate the visual quality of the images generated from the same prompts and seeds across all models. Each participant was shown 25 pairs of images from which he had to choose which of them had greater visual quality. The respondent also had the option of “equal quality” in case he could not make a clear choice. Each participant could complete the comparison only once. Of the 25 pairs shown, 3 questions were verification questions and had a clear leader in visual quality. The answers of participants who failed at least one verification question were excluded from the calculation of the results. Comparisons were allowed only in full-screen mode and only through one of the allowed browsers. Before completing the comparison, each participant was shown the following instructions: 

\begin{quote}
Thank you for participating in this evaluation.

In this study, you will be shown pairs of images generated by different neural networks from the same text prompt. From each pair, please select the image you believe has higher visual quality. The images may often look quite similar, so in addition to overall "aesthetic appeal," consider factors such as clarity, contrast, brightness, color saturation, and so on. Pay attention to generation defects, such as extra fingers or distorted bodies. If you cannot perceive any difference between the images, you may select "No difference."

The text prompt used to generate the images will not be shown, as this study focuses on evaluating visual quality, and not textual alignment.
Please note that the test includes verification questions! In these cases, the differences between the images will be clear, and selecting "indistinguishable quality" will not be considered a valid response.”
\end{quote}

\begin{figure*}[ht!]
\centering
   \includegraphics[width=.98\textwidth]{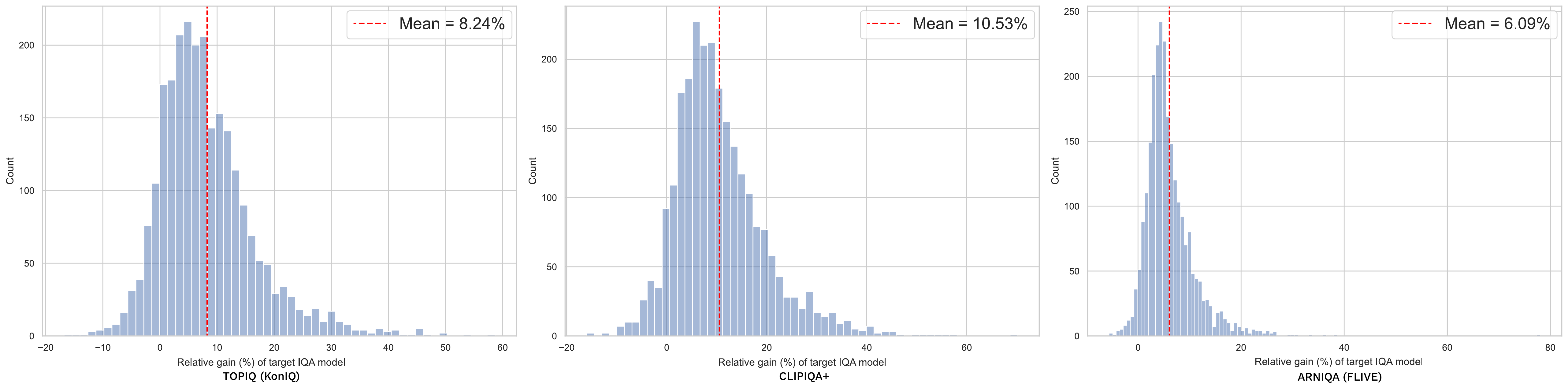}
  \caption{Distributions of relative gains defined in \ref{sec:hq_cond} across multiple generations with different seeds for IQA-Adapters trained with different IQA models. We use 25 random user-generated prompts and 100 seeds per prompt for this experiment.}
  \label{fig:sample_exp_hists}%
\end{figure*}

\section{Additional Experiments}
\label{sec:additional_exps}
\subsection{Computational Overhead}
\label{sec:comp_overhead}
\begin{table}[t]
\centering
\begin{tabular}{c|c}
\toprule
Model & Time, s \\
\midrule
Base Model (SDXL) & $3.83\pm.04$ \\
\midrule
DiffusionDPO & $3.83\pm.04$ \\
\makecell{IQA-Adapter \\w/o Separate Cross-Attn} & $3.85\pm.05$ \\
Prompt Weighting & $3.93\pm.04$ \\
IQA-Adapter & $4.07\pm.04$ \\
BeautifulPrompt & $4.15\pm.06$ \\
Q-Refine & $3.83+14.1\pm.7$ \\
\midrule
IP-Adapter & $3.99\pm.03$ \\
Ref.-based IQA-Adapter & $4.11\pm.04$ \\
StyleCrafter & $7.66\pm.08$ \\
\bottomrule
\end{tabular}

\caption{Time complexity of different generative models and conditioning methods. See Section \ref{sec:comp_overhead} for more details.}
\label{tab:time_complexity}
\end{table}

In Table \ref{tab:time_complexity}, we report time measurements for different generation methods used in this work. All evaluations were carried out in a similar environment on a single A100 80Gb GPU in float16 format and averaged across 1,000 generations. Images were generated in 1024x1024 resolution in 35 diffusion steps. We can see that the base model (SDXL) generates an image in $\sim$3.8s, and IQA-Adapter adds only $\sim$6\% to the generation time. DiffusionDPO fine-tuning method does not add any inference-time computational overhead, and Q-Refine takes triple the time of the base model to refine an \textit{already generated} image. Propmt refinement techniques generally do not add significant computational costs; however, BeatifulPrompt includes inference of a small Language Model, which adds few additional percents of computational overhead and memory use.

In the image-prompting scenario, Reference-based IQA-Adapter is a few milliseconds slower than IP-Adapter, mostly due to qualitative embedding extraction with the IQA model, and StyleCrafter is almost twice as slow as the other methods.

\subsection{Consistency across different seeds}
\label{sec:seed_consist}

To evaluate the consistency of quality improvements across different seeds, we used 25 random user-generated prompts and sampled 100 random seeds for each, resulting in 2,500 generations per model. The same set of seeds was applied to both the base model and the IQA-Adapter. Figure \ref{fig:sample_exp_hists} shows the distributions of relative gains (see Section \ref{sec:hq_cond}) across all generations for adapters trained with different IQA/IAA metrics. Positive values indicate quality improvement relative to the base model for the same seed and prompt.

The results reveal that relative gains follow a unimodal distribution with a positive mean, indicating consistent quality improvement across generations. For some occasional seeds, the base model already achieves near-optimal quality scores and leaves limited room for improvement; in these instances, the adapter introduces negligible changes, resulting in gains close to zero.

\begin{figure*}[ht!]
\centering
   \includegraphics[width=.85\textwidth]{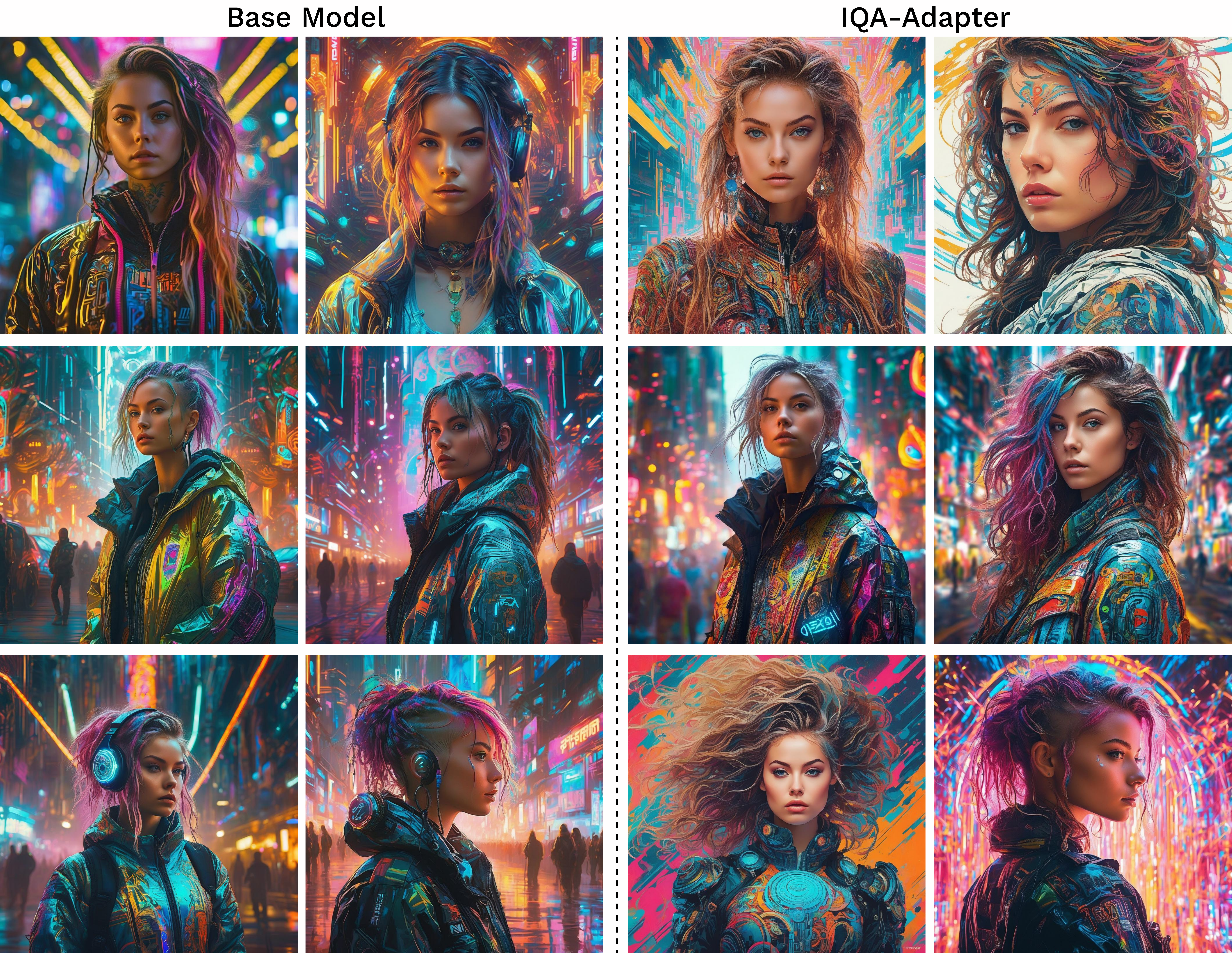}
  \caption{Examples of images generated with and without IQA-Adapter with the same prompt. The seeds are equal for corresponding images to the left and right. In this experiment, we employed the IQA-Adapter trained using the CLIP-IQA+ and LIQE-MIX models.}
  \label{fig:sample_viz}%
\end{figure*}

Figure \ref{fig:sample_viz} illustrates images generated with the same prompt and different seeds, comparing the base model to the IQA-Adapter conditioned on high quality. For this demonstration, we used a strong adapter scale ($\lambda=0.75$), which introduces noticeable stylization and detailing effects, particularly on high-frequency regions such as hair and textures.

\subsection{Generation with different input quality-conditions}
\label{sec:quality_guidance_examples_supp}
\begin{figure*}[ht!]
\centering
   \includegraphics[width=.85\textwidth]{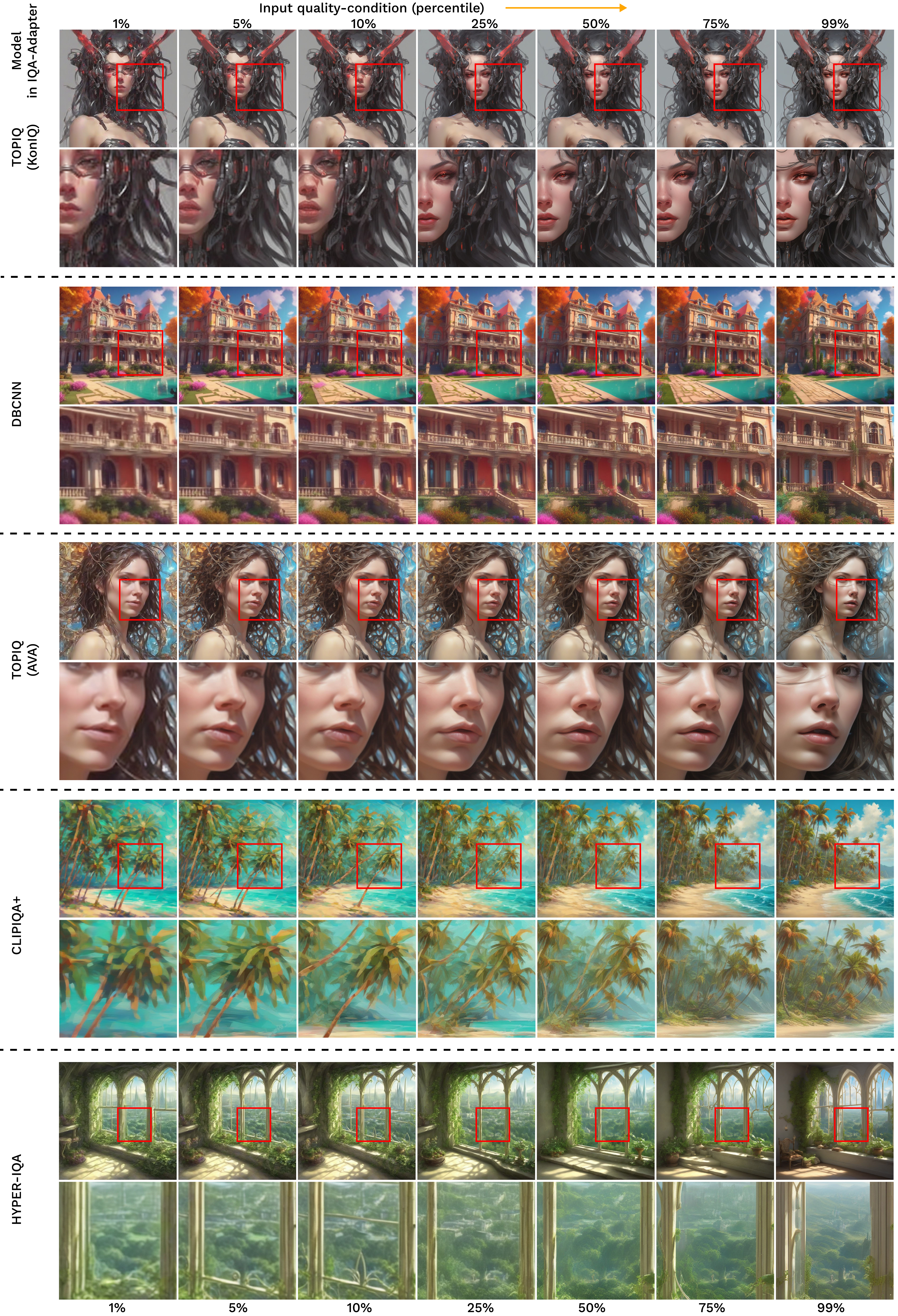}
  \caption{ Visualization of generations with different target-quality conditions with IQA-Adapters trained with different IQA/IAA models. Input quality increases from left (1-st percentile of the training set) to right (99-th percentile).}
  \label{fig:viz_quantiles}%
\end{figure*}

\begin{figure*}[ht!]
\centering
   \includegraphics[width=.75\textwidth]{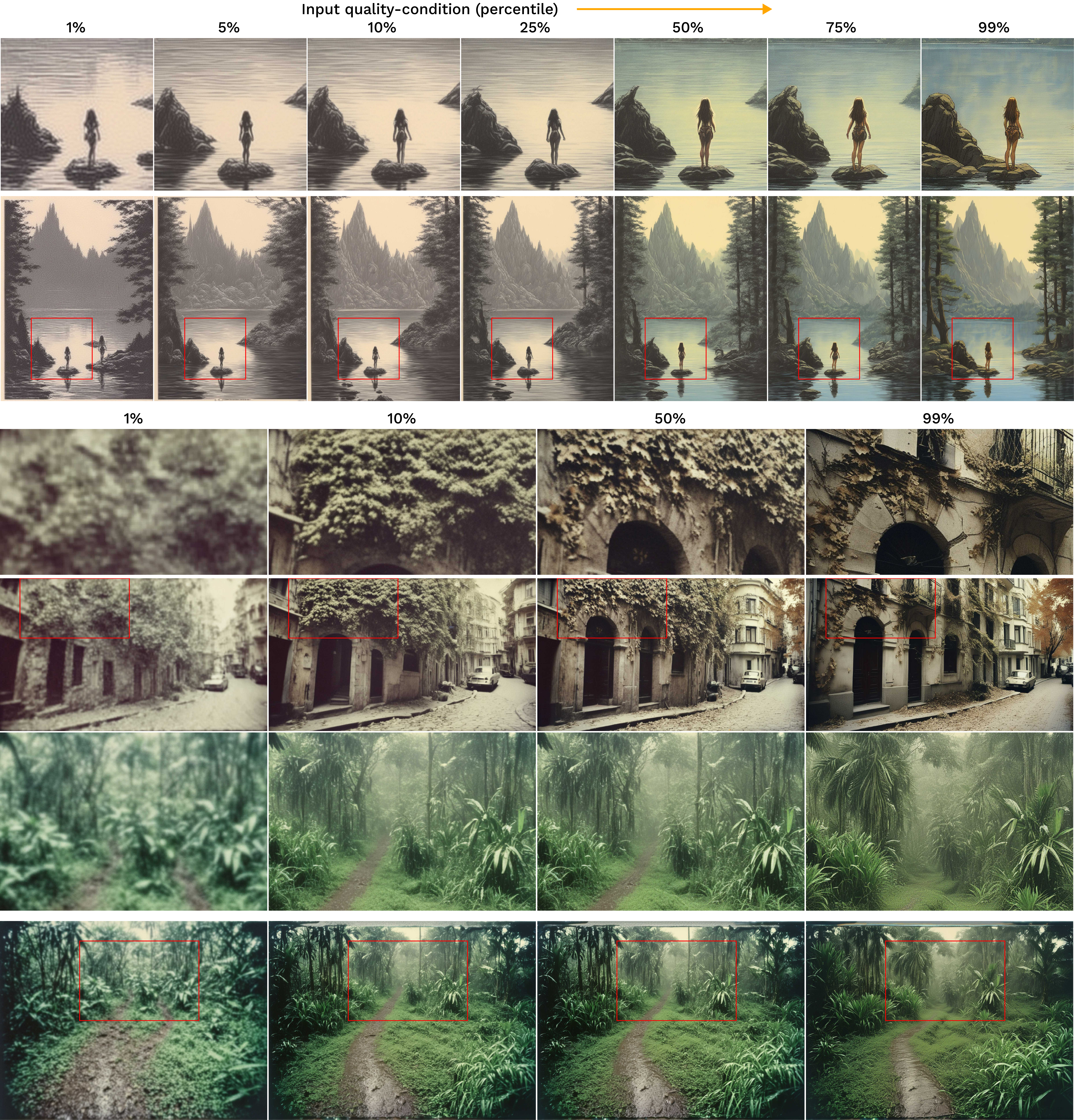}
  \caption{ Additional visualizations of IQA-Adapter quality-modulation with different aspect ratios.}
  \label{fig:viz_quantiles_new}%
\end{figure*}

Figures \ref{fig:viz_quantiles} and \ref{fig:viz_quantiles_new} illustrate the effects of modulating the IQA-Adapter with progressively higher input quality conditions. From left to right, the target quality corresponds to increasing percentiles (1st to 99th) of the target model's scores on the training dataset. Different lines represent different IQA models used during adapter training. As the target quality increases, the generated images exhibit enhanced detail and clarity, demonstrating the adapter's ability to shift image quality in alignment with the specified condition.


\section{Quality-conditioning and Adversarial Robustness of IQA models}
\label{sec:adv_patterns_supp}
Figure \ref{fig:adv_examples} presents a comparison of images generated by the base model (left column), the gradient-based method (middle column), and the IQA-Adapter (right column), alongside GradCAM visualizations of the target IQA model used for both gradient-based guidance and IQA-Adapter training. The gradient-based method often introduces artifacts that significantly alter the attention maps of the target model, inflating the quality score by exploiting architectural vulnerabilities. For instance, with the TOPIQ model (first row), new 'adversarial' objects are added to the image, capturing the model's attention and artificially boosting its scores. For TRES, grid-like patterns are generated that divert the model's focus away from the adversarial region. Similarly, with NIMA and HYPER-IQA, the method saturates the image with high-frequency details and color variations, dispersing the model’s focus. 

In contrast, the IQA-Adapter effectively preserves the target model’s saliency maps, maintaining focus on relevant objects in the scene, even when the image undergoes structural modifications.

In summary, these findings underscore the potential negative impact of direct quality optimization, which can lead to the exploitation of the target quality estimator. Gradient backpropagation through the assessor model, either at inference time or during training (e.g., through the critic model in Reinforcement Learning-based approaches),  can potentially exploit internal architectural vulnerabilities of the model. This makes the development of adversarially robust assessment models an important vector of future research. 

\begin{figure*}[ht!]
\centering
   \includegraphics[width=.87\textwidth]{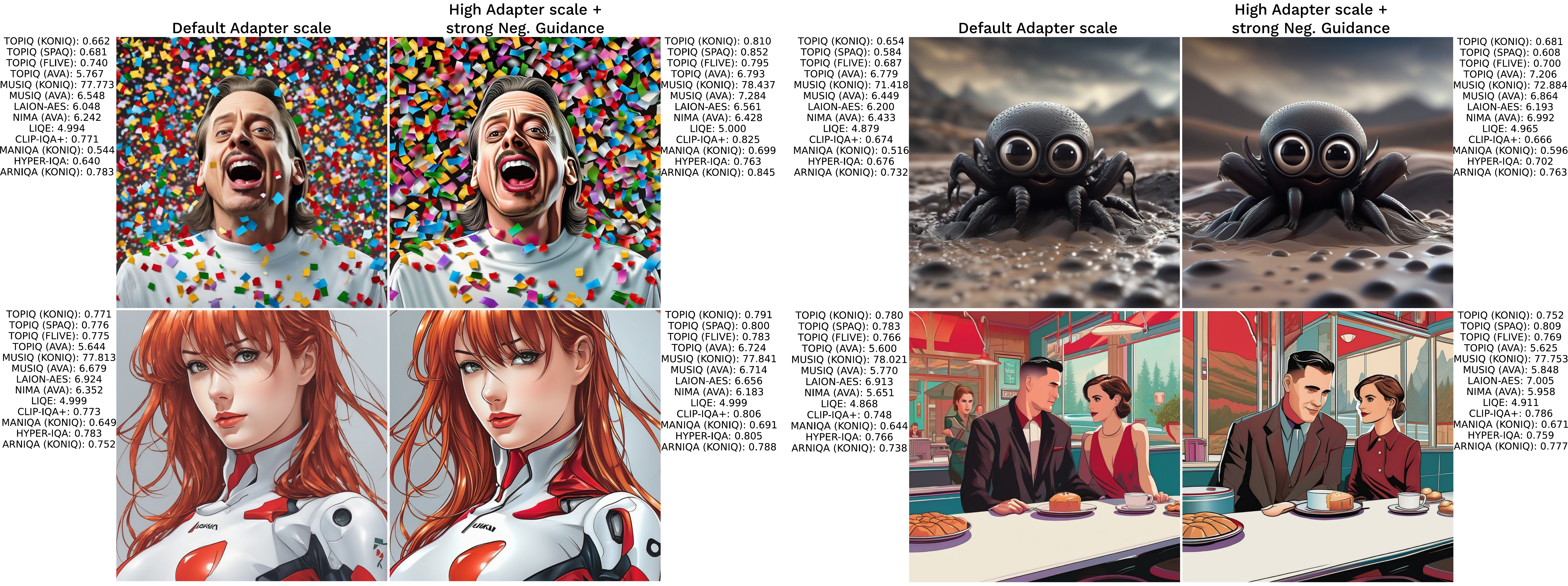}
  \caption{ Example of images generated with and without strong negative guidance ($\delta=1$) defined in Section \ref{sec:arch} under high adaptive scale ($\lambda=1$). Negative guidance magnifies the impact of the IQA-Adapter and occasionally results in the “over-stylisation” effect that is highly rated by most IQA/IAA models but usually does not reflect real quality improvement.}
  \label{fig:neg_guidance_example_supp}%
\end{figure*}

IQA-Adapter largely avoids this problem by learning qualitative features across the entire quality spectrum during training instead of focusing on the optimizationtion of quality. However, we have also found out that under excessively large adapter scale ($\lambda \geq 1$) and strong negative guidance, IQA-Adapter can sometimes produce “over-stylized” images that are highly rated by many IQA/IAA models (Figure \ref{fig:neg_guidance_example_supp}). This might indicate that the adapter identified qualitative preferences that are shared across multiple assessment models trained on different datasets and was able to exploit them.

\section{Reference-based IQA-Adapter: more visualizations}
\label{sec:ref_adapter_supp}
\begin{figure*}[ht!]
\centering
   \includegraphics[width=.90\textwidth]{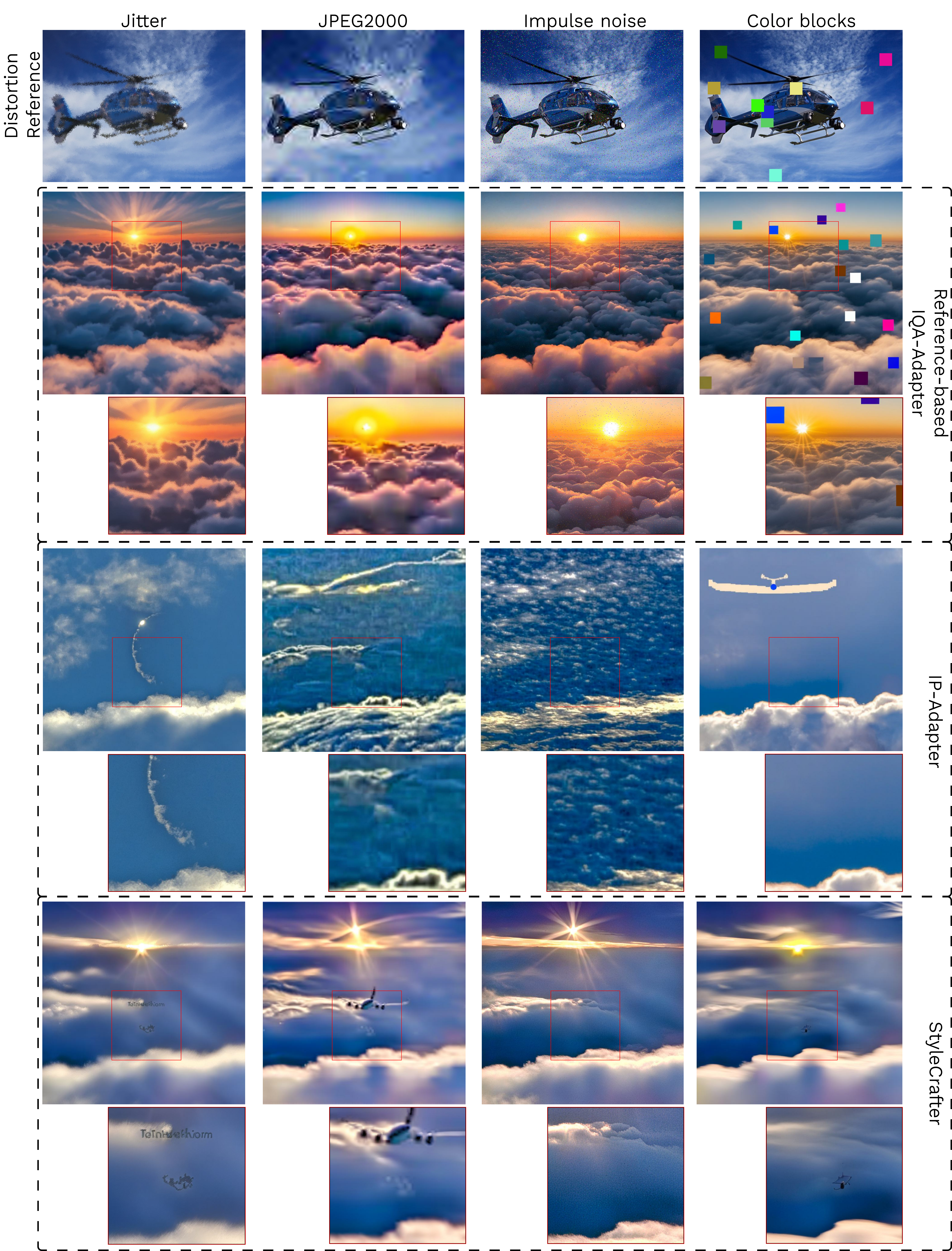}
  \caption{Text-to-Image generation with qualitative reference. First row denotes generations with Reference-based IQA-Adapter and corresponding distortion reference, second --- with IP-Adapter, and the last --- with StyleCrafter adapter. Textual prompt for all generations: \textit{"the sun rises over the clouds in the sky"}.}
  \label{fig:t2i_viz}%
\end{figure*}

Figure \ref{fig:i2i_extended} demonstrates the comparison of Reference-based IQA-Adapter and IP-Adapter in image editing task. Figure \ref{fig:t2i_viz} shows the results on Text-to-Image generation task with similar distortion references. It can be seen that other adapters copy objects and color palettes from the reference images and often fail to reproduce the distortion. We also note that we do not present the results of StyleCrafter in image editing since the official implementation of the adapter does not support SDXL Image-to-Image generation pipeline.

\begin{figure*}[ht!]
\centering
   \includegraphics[width=.90\textwidth]{imgs/new_imgs/compr/ablation_viz_compr.pdf}
  \caption{Ablation experiment: generations with IQA-Adapter with Neg. guidance enabled (1st row), without Neg. guidance (2nd row), and with a simplified IQA-Adapter without the Separate Qualitative Attention (3rd row). Simplified adapter exhibits poorer alignment with quality-condition and stronger content changes under different qualitative control signals. Negative guidance strengthens the effect of IQA-Adapter and magnifies the difference between low and high quality-conditions without significant content changes. Prompt: \textit{'A beautiful house in the woods'}.}
  \label{fig:viz_ablation}%
\end{figure*}

\begin{figure*}[ht!]
\centering
   \includegraphics[width=.85\textwidth]{imgs/new_old_imgs/supp/artifacts_supp.pdf}
  \caption{The comparison of adversarial examples generated with the gradient-based method (middle column) alongside outputs from the base model (left column) and the IQA-Adapter (right column), accompanied by their corresponding quality scores. 
  Different rows represent different target IQA/IAA models in the gradient-based method and IQA-Adapter. Even-numbered rows display GradCAM visualizations of the target IQA model applied to the images in the respective columns. The prompts are taken from the PartiPrompts dataset.}
  \label{fig:adv_examples}%
\end{figure*}

\begin{figure*}[ht!]
\centering
\hspace*{-1.0cm}
   \includegraphics[width=1.1\textwidth]{imgs/new_imgs/comparison_full.pdf}
  \caption{Comparison of different generation quality improvement methods.}
  \label{fig:comparison}%
\end{figure*}

\begin{figure*}[ht!]
\centering
   \includegraphics[width=.90\textwidth]{imgs/new_imgs/compr/i2i_extended_compr.pdf}
  \caption{Reference-based Image Editing with SDEdit using a diffusion model equipped with Reference-based IQA-Adapter and IP-Adapter. IQA-Adapter transfers qualitative information more accurately, while IP-Adapter captures the semantics of the reference image.}
  \label{fig:i2i_extended}%
\end{figure*}

\end{document}